\documentclass[sigconf,nonacm]{acmart} 
\setcopyright{none}                      
\acmDOI{} \acmISBN{} \acmConference[]{}{}{} \acmBooktitle{} \acmPrice{}
\AtBeginDocument{%
  }

\usepackage{url}            
\usepackage{booktabs}       
\usepackage{amsfonts}       
\usepackage{nicefrac}       
\usepackage{microtype}      
\usepackage{xcolor}         
\usepackage{algorithm,algpseudocode}
\usepackage{epigraph}
\usepackage{graphicx}
\usepackage{gradient-text}
\usepackage{csquotes}
\usepackage[most]{tcolorbox}
\usepackage{soul}    
\usepackage{tabularx}
\usepackage{rotating}
\usepackage[table]{xcolor}

\usepackage{xspace}     
\definecolor{olivegreen}{RGB}{107, 142, 35}
\usepackage{multirow}
\usepackage{mdframed}
\definecolor{bestblue}{RGB}{197,217,241}    
\definecolor{secondblue}{RGB}{221,235,247}  

\newcommand{\best}[1]{\cellcolor{bestblue}{\textbf{#1}}}
\newcommand{\second}[1]{\cellcolor{secondblue}{\underline{#1}}}
\usepackage{todonotes}

\newcommand{\gender}{\textcolor{pink}{\textbf{gender}}}

\newcommand{\gendered}{\textcolor{pink}{\textbf{gendered }}}
\newcommand{\Gender}{\textcolor{pink}{\textbf{Gender  }}}
\newcommand{\race}{\textcolor{red}{\textbf{race}}}  %
\newcommand{\Race}{\textcolor{red}{\textbf{Race}}}  %
\newcommand{\profession}{\textcolor{olivegreen}{\textbf{profession}}}
\newcommand{\professions}{\textcolor{olivegreen}{\textbf{professions}}}
\newcommand{\Profession}{\textcolor{olivegreen}{\textbf{Profession}}}

\begin{document}

\title{\textsc{\textsc{BiasMap}}: Leveraging Cross-Attentions to Discover and Mitigate Hidden Social Biases in Text-to-Image Generation}

\author{Rajatsubhra Chakraborty\textsuperscript{*}}
\email{rchakra6@charlotte.edu}
\affiliation{%
  \institution{University of North Carolina at Charlotte}
  \city{Charlotte}
  \state{NC}
  \country{USA}
}

\author{Xujun Che\textsuperscript{*}}
\email{xche@charlotte.edu}
\affiliation{%
  \institution{University of North Carolina at Charlotte}
  \city{Charlotte}
  \state{NC}
  \country{USA}
}

\author{Depeng Xu}
\email{dxu7@charlotte.edu}
\affiliation{%
  \institution{University of North Carolina at Charlotte}
  \city{Charlotte}
  \state{NC}
  \country{USA}
}

\author{Cori Faklaris}
\email{cfaklari@charlotte.edu}
\affiliation{%
  \institution{University of North Carolina at Charlotte}
  \city{Charlotte}
  \state{NC}
  \country{USA}
}

\author{Xi Niu}
\email{xniu2@charlotte.edu}
\affiliation{%
  \institution{University of North Carolina at Charlotte}
  \city{Charlotte}
  \state{NC}
  \country{USA}
}

\author{Shuhan Yuan}
\email{Shuhan.Yuan@usu.edu}
\affiliation{%
  \institution{Utah State University}
  \city{Logan}
  \state{UT}
  \country{USA}
}



\begin{abstract}
Bias discovery is critical for black-box generative models, especially text-to-image (TTI) models. Existing works predominantly focus on output-level demographic distributions, which do not necessarily guarantee concept representations to be disentangled post-mitigation. We propose \textsc{BiasMap}, a model-agnostic framework for uncovering latent concept-level representational biases in stable diffusion models. \textsc{BiasMap} leverages cross-attention attribution maps to reveal structural entanglements between demographics (e.g., gender, race) and semantics (e.g., professions), going deeper into representational bias during the image generation. Using attribution maps of these concepts, we quantify the spatial demographics-semantics concept entanglement via Intersection over Union (IoU), offering a lens into bias that remains hidden in existing fairness discovery approaches. In addition, we further utilize \textsc{BiasMap} for bias mitigation through energy-guided diffusion sampling that directly modifies latent noise space and minimizes the expected SoftIoU during the denoising process. Our findings show that existing fairness interventions may reduce the output distributional gap but often fail to disentangle concept-level coupling, whereas our mitigation method can mitigate concept entanglement in image generation while complementing distributional bias mitigation.
\end{abstract}



\keywords{text-to-image generation, social bias, bias mitigation}


\maketitle
\begingroup
  \renewcommand\thefootnote{}
  \footnotetext{\textsuperscript{*}Equal contribution.}
\endgroup
\section{Introduction}
\label{sec:intro}
\enquote{\textbf{\textit{With great power, comes great responsibility.}}} While the iconic quote from Spider-Man's Uncle Ben Parker was meant for superheroes, it also applies to generative models. Stable Diffusion (SD) models \cite{rombach2022high, podellsdxl, esser2024scaling} hold significant power in creating highly realistic images from input text. However, much like Spider-Man's webs, their outputs are often entangled with inherent biases \cite{vazquez2024taxonomy} that frequently go unnoticed. Recent SD models achieve their impressive capabilities by learning statistical patterns from massive internet-sourced datasets comprising billions of images and captions \cite{schuhmann2022laion5bopenlargescaledataset}. Yet, these datasets inherently reflect societal biases \cite{birhane2024dark, fabbrizzi2022survey}, implicitly perpetuating or amplifying stereotypes involving sensitive demographic attributes such as \gender{} and \race{}. Such biases pose significant ethical and fairness challenges. Moreover, due to SD's internal opacity \cite{shi2025dissecting}, precisely identifying or addressing biases at a representational level remains difficult. Generally, biases originate from two primary sources: data-level imbalances in training sets \cite{seshadri2023biasamplificationparadoxtexttoimage, luccioni2023stablebiasanalyzingsocietal} and latent representational entanglements \cite{zhu2025latentexplainerexplaininglatentrepresentations} which are hidden internal correlations between demographic (e.g., \gender{}, \race{}) and semantics (e.g., \professions{}) learned implicitly. Even when data-level issues are corrected, latent entanglements often persist, subtly embedding stereotypes conceptually \cite{locatello2019fairness}. We explicitly define {latent entanglement} as representational overlaps between demographics and semantics, signifying implicit associations that remain independent of explicit textual conditioning.

Prior works \cite{aldahoul2024ai, luccioni2023stable} have primarily focused on output-level observation in SD, examining skews in demographic distributions at the generated image level. While these approaches provide valuable insights, they offer limited understanding of the internal representational structures that underpin these biases. Recently, some efforts \cite{mandal2024generated, kim2024unlocking} have been made to inspect biases within the diffusion process itself. However, these studies lack fine-grained spatial-level indicators, offering no direct method to identify precisely which regions or pixels are impacted by bias or how deeply demographics become entangled with semantics.
To overcome these limitations and move beyond superficial, output-level bias audits, it is essential to inspect and quantify the latent representational entanglement spatially within generative models. A deeper understanding of how demographics intertwine internally with semantics would enable more targeted and effective bias mitigation interventions, going beyond merely adjusting output distributions to structurally addressing biases at the representational level. Therefore, our work explicitly targets this crucial research gap and poses the following central research questions:

\noindent \fbox{%
  \parbox{\dimexpr\linewidth-2\fboxsep-2\fboxrule}{%
    \noindent\textbf{RQ1:} \textit{How can we leverage attribution mapping to explain the source of bias for generation in SD?}
  }%
}

\noindent \fbox{%
  \parbox{\dimexpr\linewidth-2\fboxsep-2\fboxrule}{%
    \noindent\textbf{RQ2:} \textit{How do we quantify the bias in SD in the form of demographics-semantics concept entanglement?}
  }%
}

\noindent \fbox{%
  \parbox{\dimexpr\linewidth-2\fboxsep-2\fboxrule}{%
    \noindent\textbf{RQ3:} \textit{How do we disentangle demographics and semantics in SD to achieve fair generation?}
  }%
  
}

\noindent To address these questions, we propose a model-agnostic framework, called \textsc{\textbf{\textsc{BiasMap}}}. It first utilizes cross-attention attribution maps to quantify latent representational entanglements in text-to-image diffusion models. It then further mitigates biased concept entanglement through energy-guided diffusion sampling. Our primary contributions are:

\begin{itemize}
    \item A novel bias localization method that precisely identifies and quantifies representational entanglement between demographic and semantic concepts.

    \item The introduction of Intersection-over-Union (IoU) as a metric for effectively quantifying demographics-concept biases, complementing traditional distribution-based metrics such as Risk Difference (RD).

    \item A debiasing method through energy-guided diffusion sampling that directly modifies latent noise space and minimizes the expected SoftIoU during the diffusion.

    \item Empirical evidence demonstrating that \textsc{BiasMap} provides deeper insights into latent representational biases and instructs bias mitigation in the denoising process.
\end{itemize}



\section{Related Works}
\label{sec: related_works}
\paragraph{Image Synthesis and TTI models.}
Early image synthesis relied on deterministic algorithms and feature engineering \cite{efros1999texture,heeger1995pyramid}, which limited realism and flexibility. Deep learning \cite{lecun2015deep} introduced Variational Autoencoders (VAEs) \cite{kingma2013auto} and Generative Adversarial Networks (GANs) \cite{goodfellow2020generative}, improving generation quality despite VAEs producing blurry images and GANs suffering from training instability. Early text-to-image (TTI) approaches used VAEs with text sequences \cite{mansimov2015generating}. The emergence of diffusion models \cite{sohl2015deep} treated generation as a denoising process from pure noise. Latent Diffusion Models operated efficiently in compressed latent space, reducing computational costs while maintaining image fidelity. Stable Diffusion \cite{rombach2021highresolution} became the foundational TTI model, evolving from 512×512 images with CLIP encoders to SDXL \cite{podellsdxl} with 3.5B parameters and 1024×1024 support, and eventually to recent variants \cite{esser2024scaling} with up to 8B parameters.
OpenAI's DALL-E \cite{ramesh2021zero} used transformer architecture with discrete variational autoencoders (dVAE). DALL-E 2 \cite{ramesh2022hierarchical} improved this with a two-stage framework: generating CLIP image embeddings from text, then decoding images from these embeddings, significantly enhancing semantic understanding and generation quality. Google's Imagen \cite{saharia2022photorealistic} employed pretrained text encoders to condition cascaded diffusion models, achieving photorealistic images with nuanced language understanding. Google's Parti \cite{yu2022scaling} approached TTI as sequence-to-sequence generation, using autoregressive transformers to generate image token sequences from text, enabling complex compositions with extensive world knowledge integration.\paragraph{Interpretability and Bias Discovery in SD} 
Recent studies have focused on elucidating the internal mechanisms of diffusion models, particularly within the SD family, to understand generation processes and discover internal biases in TTI models. Diffusion Attentive Attribution Maps (DAAM) \cite{tang2023daam} generates pixel-level attribution maps by upscaling and aggregating cross-attention scores from SD's denoising network. Diffusion Lens \cite{toker2024diffusion} examines text encoder components by generating images conditioned on intermediate text representations, revealing how textual information is processed during synthesis, though it primarily focuses on text encoders in isolation, potentially missing text-image interactions. Open Vocabulary Attention Maps (OVAM) \cite{marcos2024open} provides a training-free method for generating attention maps for arbitrary words beyond original prompts, incorporating lightweight token optimization for enhanced accuracy.

Recent bias discovery methods \cite{li2024self, seshadri2023bias, wang2023t2iat, shi2025dissecting} focus on intermediate representational observations to understand inherent model biases. Li et al. \cite{li2024self} proposed self-supervised techniques for extracting interpretable latent directions corresponding to semantic attributes, enabling attribute disentanglement and bias axis discovery, though limited to binary attributes. Seshadri et al. \cite{seshadri2023bias} introduced a bias amplification paradox framework comparing generated image attributes against training caption distributions, revealing that SD amplifies biases even with neutral prompts due to training data priors and prompt alignment mismatches. OpenBias \cite{d2024openbias} developed a flexible pipeline for open-set bias discovery using image generation, vision-language models, and question-answering modules to identify both known and emergent biases across diverse prompts. Wu et al. \cite{wu2025revealing} demonstrated that gender associations influence not only face and body generation but also object placement and compositional structure, indicating entrenched priors in both text encoders and image generators. The most recent work by Shi et al. \cite{shi2025dissecting} uncovered localized generative structures responsible for encoding bias-correlated concepts, proposing patching interventions for bias-aware control without architectural retraining.
\paragraph{Bias Mitigation.} 
Existing bias mitigation approaches for diffusion models primarily target output-level demographic distributions. Fair Diffusion \cite{friedrich2023fairdiffusioninstructingtexttoimage} allows users to guide model outputs via human instructions to achieve desired demographic representations through explicit prompt engineering. Inclusive Text-to-Image Generation (ITI-GEN) \cite{zhang2023inclusive} uses reference images to guide generation and ensure diverse attribute inclusion without model fine-tuning. Text-to-Image Model Editing (TIME) \cite{orgad2023editing} modifies implicit assumptions by updating cross-attention layers based on source and destination prompts, while research in \cite{parihar2024balancing} introduced distribution guidance to condition the reverse diffusion process on sensitive attribute distributions during sampling.

Recent advances have explored training-free approaches and complex bias scenarios. Research in \cite{kim2025rethinking} developed "weak guidance" that exploits Stable Diffusion's potential to reduce bias without additional training by guiding random noise toward clustered "minority regions" while preserving semantic integrity. Research in \cite{yesiltepe2024mist} introduced MIST for addressing intersectional bias by modifying cross-attention maps in a disentangled manner to tackle biases at the intersection of multiple social identities. Additionally, research in \cite{park2025fair} proposed Entanglement-Free Attention (EFA) to address attribute entanglement, where bias adjustments to target attributes unintentionally alter non-target attributes, achieving fair target attribute distribution while preserving non-target attributes and maintaining generation capabilities.

\section{\textsc{BiasMap}}
\label{sec:biasmap}
\subsection{Preliminary}
\label{subsec: prelim}

Let \( \boldsymbol{\epsilon}\) denote a stable diffusion model. Given a prompt \( P\) and noise \( \mathbf{z}\), the model generates the corresponding image $\mathbf{I} = \boldsymbol{\epsilon}_\theta(\mathbf z,\,P)$ with shape $W \times H \times C$. 

In generative TTI models, cross-attention integrates text into image synthesis. 
OVAM attributes spatial influence to arbitrary concepts, even those absent from input prompts. For an arbitrary concept $a$, which does not need to be in the original prompt $P$ used to generate the image $\mathbf{I}$, OVAM generates an attention attribution map ${\mathbf{M}}_a(\mathbf{I})$ to interpret the spatial region related to the concept $a$. 
To construct OVAM, the attribution prompt $P'$ with $a \in P'$  is converted by CLIP encoder as ${\mathbf{X}}' \in \mathbb{R}^{d_E \times d_{{\mathbf{X}}'}}$ , where $d_E$ is the embedding dimension and $d_{{\mathbf{X}}'}$ is the number of tokens.
Without loss of generosity, the concept $a$ is expressed as a single token $a$.
For generating the open-vocabulary attention matrices for even concept $a\notin P$, OVAM uses \( \ell^{(i)}_K \) as key projection at each block \( i \) to compute the attribution keys: \( {\mathbf{K}}'_i = \ell^{(i)}_K({\mathbf{X}}') \). 
During denoising, pixel-space queries are extracted at block \( i \), timestep \( t \): \( {\mathbf{Q}}_{i,t} = \ell^{(i)}_Q({\mathbf{h}}_{i,t}) \)
, where \( {\mathbf{h}}_{i,t} \) is the $i$-th convolutional block output at time step $t$ and \( \ell^{(i)}_Q  \) is learned projection at block $i$.
The cross-attention matrix ${\mathbf{A}} \in \mathbb{R}^{W^{(i)}\times H^{(i)} \times d_H^{(i)} \times d_{{\mathbf{X}}'}}$ is computed for each block $i$ and time step $t$:
\begin{equation}
{\mathbf{A}}({\mathbf{Q}}_{i,t}, {\mathbf{K}}'_i) = \text{softmax}\left( \frac{{\mathbf{Q}}_{i,t} {\mathbf{K}}_i'^\top}{\sqrt{d}} \right),
\label{eq:att_weights}
\end{equation}
where \( d \) is the query/key dimensionality, $W^{(i)} \times H^{(i)}$ is the reduced latent space shape at block $i$, $d_H^{(i)}$ is the number of attention heads at block $i$. 

To generate the attribution map ${\mathbf{M}}_a(\mathbf{I})$, OVAM aggregates the matrices across blocks, timestamps, and attention heads for the slices associated with token $a$:
\begin{equation}
{\mathbf{M}}_a(\mathbf{I}) = \sum_{i,t,l} \text{resize}\left( \mathbf{A}_{l,a}({\mathbf{Q}}_{i,t}, {\mathbf{K}}'_i) \right) \in \mathbb{R}^{W \times H},
\label{eq:ovam}
\end{equation}
where \( \mathbf{A}_{h,k} \) refers to the slice associated with the $l$-th attention head and token $a$,
and \(\text{resize}(\cdot)\) normalizes resolution by bilinear interpolation. 
The map ${\mathbf{M}}_a(\mathbf{I})  \in \mathbb{R}^{W \times H}$ localizes token influence for concept probing.
When both $P$ and $P'$ are identical, the heatmaps are equivalent to directly extracting and aggregating the cross-attention matrices computed during image synthesis.

\subsection{Bias Discovery via Attribution Maps}
\label{sec:discovery}
\begin{figure}[t]  
  \centering
  \includegraphics[width=\linewidth]{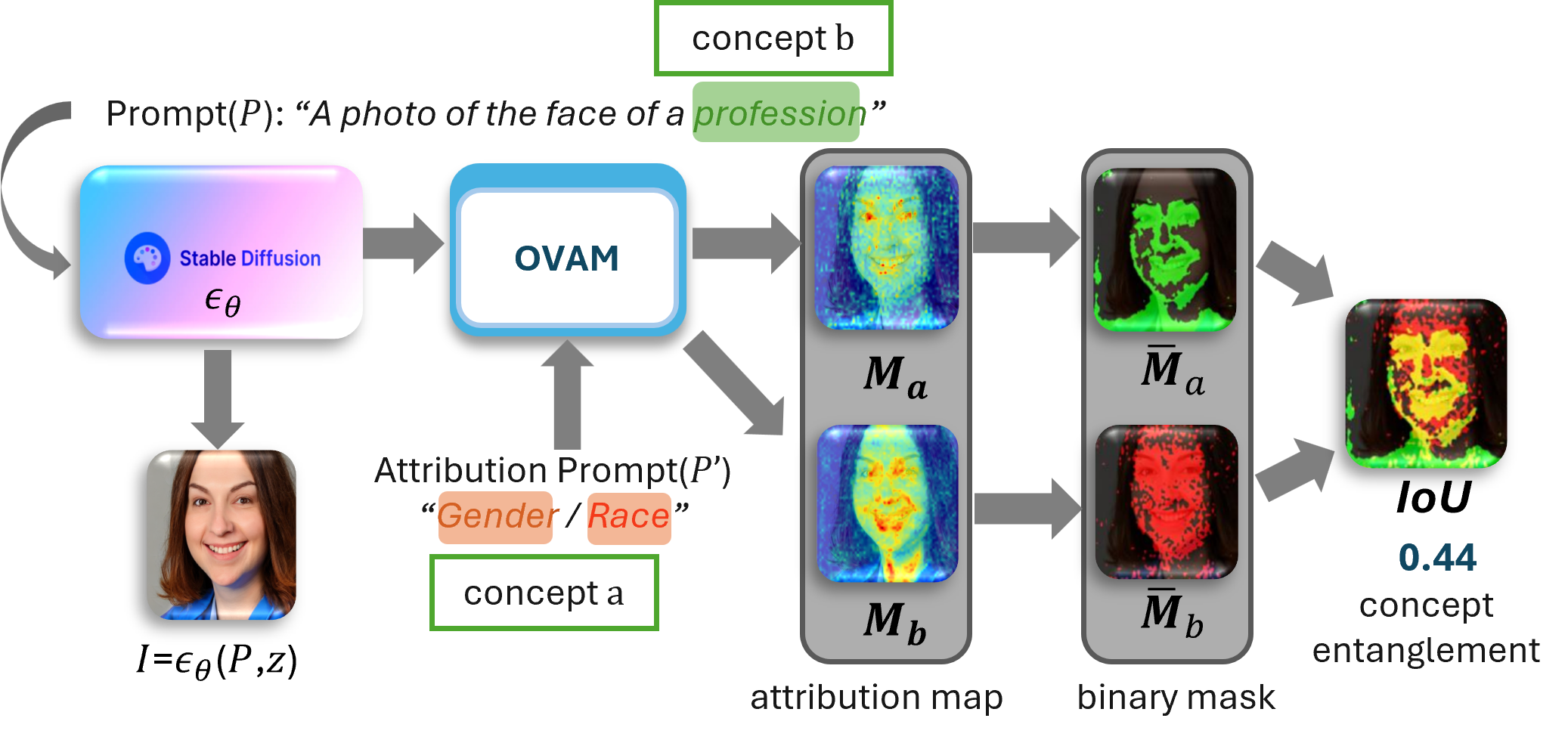}
  \caption{Bias Discovery via Attribution Maps Pipeline.}%
  \Description{Visualization of the detection process.}
  \label{fig:detection}
\end{figure}

To interpret biased concept association, we propose \textsc{\textbf{\textsc{BiasMap}}}, which spatially localizes concept entanglement during image generation via concept attribution maps, as seen in Figure \ref{fig:detection}.

\noindent\textbf{Bias Localization.}
To evaluate the generation of $\mathbf{I} = \boldsymbol{\epsilon}_\theta(\mathbf z,\,P)$, we define two attribution prompts with embeddings \( P'_a \) and \( P'_b \) containing concepts $a$ and $b$, respectively.
In the context of bias discovery, concept $a$ denotes the \textbf{demographics} (e.g., \gender{} or \race{}) and concept $b$ denotes the \textbf{semantics} (e.g., \profession{}). In a common TTI setting explored in previous works, $a \notin P$ and $b \in P$.
Using Eq.~\ref{eq:ovam}, we compute the aggregated attribution maps \( {\mathbf{M}}_a \) and \( {\mathbf{M}}_b \) indicating spatial attribution for each concept. We model \textbf{concept entanglement} as the similarity of cross-attention attribution maps in the pixel space between two concepts.
More specifically, we focus on the attribution maps of the high cross-attention regions in the pixel space. We localize the high attention regions with at a threshold quantile $q$ and generate binary masks:
\begin{equation}
\label{eq:quantile_thresholding}
\bar{\mathbf{M}}_a[x, y] =
\begin{cases}
1, & \text{if } {\mathbf{M}}_a[x,y] \geq \operatorname{Quantile}_q\bigl({\mathbf{M}}_a\bigr), \\[6pt]
0, & \text{otherwise.}
\end{cases}
\end{equation}
where \( [x, y] \) denotes the 2D spatial coordinates in the attention heatmap of resolution \( W \times H \), and ${\mathbf{M}}_a[x,y] \in \{0, 1\}$. 
This yields binary masks \( \bar{\mathbf{M}}_a \) and \( \bar{\mathbf{M}}_b \) representing regions most influenced by the respective concepts.

We compute the \textbf{Intersection over Union (IoU)} between these masks to quantify entanglement:
\begin{align}
\mathrm{IoU}(\bar{\mathbf{M}}_a, \bar{\mathbf{M}}_b) 
= \frac{ \sum_{x,y} \bar{\mathbf{M}}_a[x,y] \cdot \bar{\mathbf{M}}_b[x,y] }
{ \sum_{x,y} \max(\bar{\mathbf{M}}_a[x,y], \bar{\mathbf{M}}_b[x,y]) }.
\label{eq:iou}
\end{align}
\begin{tcolorbox}[
    colframe=green!40,
    colback=green!10,
    boxrule=0.5pt,
    arc=2mm,
    left=1mm,
    right=1mm,
    top=1mm,
    bottom=1mm,
    title=\textbf{Intuition},
    fonttitle=\small\bfseries,
    coltitle=black
]
\small
If \textbf{concept entanglement} exists between demographics and semantics, the attention maps should have \textbf{substantial intersection} over spatial regions, i.e., the same pixels are influenced by both concepts during generation.
\end{tcolorbox}
\noindent Lower IoU indicates better separation of demographics and semantics concepts in image generation.

\noindent\textbf{Block-wise Bias Localization.}
Eq.~\ref{eq:ovam} shows the aggregated attribution map across all blocks. Since concepts are generated in different blocks, we further dive into the block-wise attribution map at block~$i$:
\begin{equation}
{\mathbf{M}}_a^{(i)}(\mathbf{I}) = \sum_{t,l} \text{resize}\left( \mathbf{A}_{l,a}({\mathbf{Q}}_{i,t}, {\mathbf{K}}'_i) \right) \in \mathbb{R}^{W \times H}.
\label{eq:ovam_block}
\end{equation}
Similarly, we obtain the block-wise binary masks $\bar{\mathbf{M}}_a^{(i)}$ and $\bar{\mathbf{M}}_b^{(i)}$ for two concepts and compute the \textbf{Block-wise Intersection over Union (BIoU)} to analyze at what depth entanglement occurs.
\begin{align}
\mathrm{BIoU}^{(i)}(\bar{\mathbf{M}}_a^{(i)}, \bar{\mathbf{M}}_b^{(i)}) 
= \frac{ \sum_{x,y} \bar{\mathbf{M}}_a^{(i)}[x,y] \cdot \bar{\mathbf{M}}_b^{(i)}[x,y] }
{ \sum_{x,y} \max(\bar{\mathbf{M}}_a^{(i)}[x,y], \bar{\mathbf{M}}_b^{(i)}[x,y]) }.
\label{eq:iou_block}
\nonumber
\end{align}
For each individual image $\mathbf{I}$, an average BIoU over all blocks is computed.
\begin{equation}
\text{BIoU}(a, b) = \frac{1}{N} \sum_{i=1}^{N} \mathrm{BIoU}^{(i)}(\bar{\mathbf{M}}_a^{(i)}, \bar{\mathbf{M}}_b^{(i)}),
\end{equation}
where $N$ is the number of blocks.

\noindent\textbf{Difference from Risk Difference.}
Previous works only focus on group fairness in generation output distribution. The \textbf{Risk Difference (RD)}, defined as $\text{RD}(a_1, a_2) = \left| \Pr(a_1) - \Pr(a_2) \right|$, where \( a_1 \), \( a_2 \) are different demographic groups (e.g., ``male'' and ``female''), measures group-level bias through demographic parity across samples, addressing distributional fairness. 
Conversely, IoU or BIoU quantifies representational entanglement by measuring spatial co-activation patterns for each individual generation. it reveals how demographics become structurally coupled with semantics concept in the latent space. We can also extend concept entanglement to a group setting. We evaluate the mean IoU (or BIoU) for different images generated by $\boldsymbol{\epsilon}_\theta$ with the same instruction prompt $P$ but different noises $\mathbf{z}$.

\subsection{Bias Mitigation via Energy‑Guided Diffusion Sampling}
\begin{figure*}[t]   
  \centering
  \includegraphics[width=0.8\linewidth]{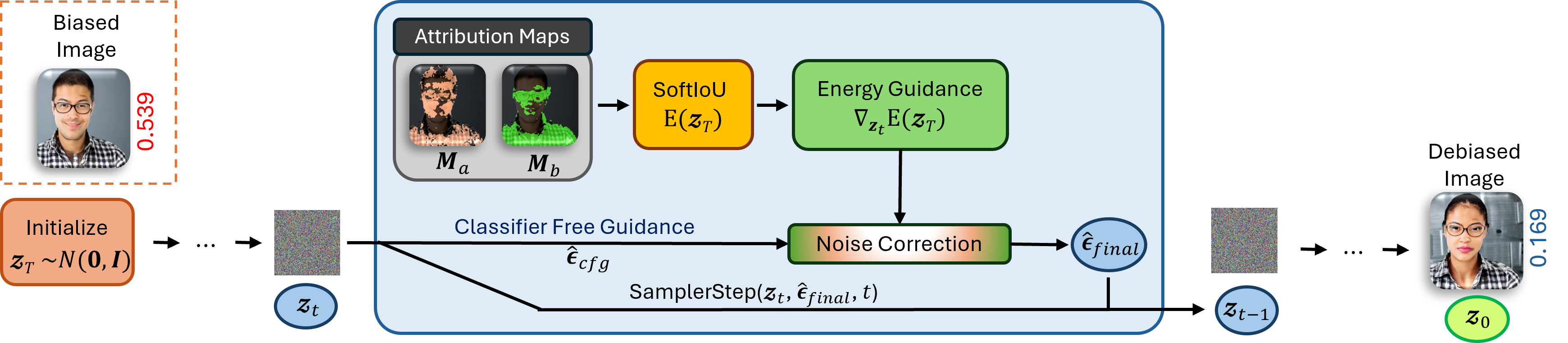}
  \caption{\textsc{BiasMap} Mitigation via Energy‑Guided Diffusion Sampling Pipeline.}
  \Description{Visualization of the mitigation process.}
  \label{fig:mitigation}
\end{figure*}

Building on the bias discovery in Section \ref{sec:discovery}, {\textsc{\textbf{BiasMap}}} now steers the diffusion sampler so that the biased concept entanglement reduces during generation.

Let $\mathbf{z}_t$ denote the latent state at diffusion timestep $t \in \{0, 1, \ldots, T\}$
The denoising model provides a noise prediction function $\boldsymbol{\epsilon}_\theta$ parameterized by $\theta$, which estimates the noise component $\boldsymbol{\epsilon} = \boldsymbol{\epsilon}_{\theta}(\mathbf{z}_t,\cdot)$ added during the forward diffusion process at each step $t$.
The fundamental challenge for bias mitigation lies in sampling from a posterior distribution that minimizes representational entanglement while preserving semantic fidelity and generative quality. 

We formulate bias mitigation as a principled energy-based guidance problem within the diffusion sampling framework, leveraging the theoretical foundations of score-based generative models and Bayesian posterior conditioning. As shown in Figure \ref{fig:mitigation}, {\textsc{\textbf{BiasMap}}} directly modifies the latent noise space~\cite{NEURIPS2021_49ad23d1} during the denoising process through a mathematically rigorous energy-based guidance framework that seeks an optimal sampling trajectory that minimizes expected SoftIoU (as the energy function).
\algrenewcommand\alglinenumber[1]{\color{red}#1}
\begin{algorithm}[H]
\caption{Energy-Guided Diffusion Sampling}
\label{alg:energy_guided_sampling}
\begin{algorithmic}[1]
\Require Diffusion model $\boldsymbol{\epsilon}_\theta$, generation prompt $P$, attribution prompt $P'$ with tokens $\{a, b\}$, CFG scale $\gamma$, Energy guidance scale $\lambda$, percentile threshold $q$.
\State Initialize $\mathbf{z}_T \sim \mathcal{N}(\mathbf{0}, \mathbf{I})$
\For{$t = T$ \textbf{downto} $1$}
    \State $\hat{\boldsymbol{\epsilon}}_{\text{un}} \leftarrow \boldsymbol{\epsilon}_\theta(\mathbf{z}_t, \emptyset)$ \Comment{Unconditional noise prediction}
    \State $\hat{\boldsymbol{\epsilon}}_{\text{con}} \leftarrow \boldsymbol{\epsilon}_\theta(\mathbf{z}_t, P)$ \Comment{Conditional noise prediction}
    \For{each token $w \in \{a, b\}$}
        \State ${\mathbf{M}}_w^{(t)} \leftarrow \sum_{i,l} \text{resize}(\mathbf{A}_{l,w}(\mathbf{Q}_{i,t}(\mathbf{z}_t), \mathbf{K}'_i))$ \Comment{OVAM}\label{line:ovam}
        \State $\tilde{{\mathbf{M}}}_w^{(t)} \leftarrow \text{SoftTopK}({\mathbf{M}}_w^{(t)}, q)$ \Comment{Differentiable soft masks} \label{line:softtopk}
    \EndFor
    \State $E(\mathbf{z}_t)\leftarrow \text{SoftIoU}(\tilde{{\mathbf{M}}}_a^{(t)}, \tilde{{\mathbf{M}}}_b^{(t)})$ \Comment{Differentiable IoU loss}\label{line:iou-loss}
    \State $\hat{\boldsymbol{\epsilon}}_{\text{cfg}} \leftarrow \hat{\boldsymbol{\epsilon}}_{\text{un}} + \gamma(\hat{\boldsymbol{\epsilon}}_{\text{con}} - \hat{\boldsymbol{\epsilon}}_{\text{un}})$ \Comment{Classifier-free guidance}\label{line:cfg}
    \State $\hat{\boldsymbol{\epsilon}}_{\text{final}} \leftarrow \hat{\boldsymbol{\epsilon}}_{\text{cfg}} + \lambda \sqrt{1 - \bar{\alpha}_t} \nabla_{\mathbf{z}_t} E(\mathbf{z}_t)$ \Comment{Energy-based guidance}\label{line:energy-based-guidance}
    \State $\mathbf{z}_{t-1} \leftarrow \text{SamplerStep}(\mathbf{z}_t, \hat{\boldsymbol{\epsilon}}_{\text{final}}, t)$ \Comment{Diffusion step with arbitrary sampler}\label{line:sampler}
\EndFor
\State \Return $\mathbf{z}_0$
\end{algorithmic}
\end{algorithm}
\subsubsection{Problem Formulation} $\xspace$\\
\textbf{Cross-Attention Attribution Maps.}
For each timestep $t$, {\textsc{\textbf{BiasMap}}} extracts cross-attention attribution maps for both demographic attribute $a$ and semantic concept $b$ using OVAM (\textcolor{red}{Line \ref{line:ovam}} in Algorithm \ref{alg:energy_guided_sampling}). Given the current latent state $\mathbf{z}_t$ and attribution prompt $P'$ containing both concepts, it computes the raw attribution maps ${\mathbf{M}}_w^{(t)}$ at each diffusion timestep $t$ for each token $w \in {a, b}$ dynamically, rather than post-generation like Eq. \ref{eq:ovam}.
The spatial queries ${Q}_{i,t}(\mathbf{z}_t)$ from the current latent state interact with textual keys ${K}'_i$ from the attribution prompt to produce attention maps that capture the real-time spatial influence of each concept during generation. 

\noindent\textbf{Mitigation Objective.} We formulate bias mitigation as an optimization over diffusion trajectories, our primary objective is to find an optimal sampling trajectory $\mathcal{Z}^* = \{\mathbf{z}_t^*\}_{t=0}^T$ that minimizes the expected concept entanglement while maintaining fidelity to the conditioning prompt $P$ that minimizes expected concept entanglement:
\begin{equation}
\begin{aligned}
\mathcal{Z}^*
  &= \arg\min_{\mathcal{Z}}
     \;\mathbb{E}_{t \sim \mathcal{U}(1,T)}
     \!\bigl[\operatorname{IoU}^{(t)}(\bar{\mathbf{M}}_a^{(t)}, \bar{\mathbf{M}}_b^{(t)})\bigr], \\
\text{s.t.}\;&
  \mathbf{z}_{t-1} \sim p_{\theta}\bigl(\mathbf{z}_{t-1}\mid\mathbf{z}_t,\,P\bigr),
  \nonumber
\end{aligned}
\end{equation}
where $\text{IoU}^{(t)}$ is computed dynamically during the sampling process rather than as a post-hoc analysis metric in Eq. \ref{eq:iou}. This formulation ensures that bias mitigation respects the underlying diffusion dynamics while systematically reducing the spatial co-activation patterns that \textsc{BiasMap} identified as the source of representational bias.

However, direct application of IoU for gradient-based optimization is prevented by the non-differentiable thresholding operations inherent in the binary mask generation process of Eq. \ref{eq:quantile_thresholding}. To address this fundamental limitation while preserving the semantic meaning of the IoU metric, we develop a differentiable relaxation based on entropy-regularized optimal transport theory. This is operationalized through the SoftTopK function in \textcolor{red}{Line \ref{line:softtopk}} of Algorithm \ref{alg:energy_guided_sampling}, which transforms the discrete attention maps from bias discovery into differentiable soft masks suitable for real-time guidance.

\subsubsection{Energy-based Guidance Framework} $\xspace$\\
\textbf{Energy Function Construction.} To enable gradient-based optimization, we construct an auxiliary energy function $E$ that serves as a differentiable surrogate for concept entanglement.
The key innovation lies in extending the static post-generation IoU analysis from Section \ref{sec:discovery} to a dynamic, differentiable energy function that can guide the sampling process in real-time. Specifically, the energy function is instantiated as the SoftIoU operation shown on \textcolor{red}{Line \ref{line:iou-loss}} of Algorithm \ref{alg:energy_guided_sampling}:
\begin{equation}
E(\mathbf{z}_t) = \text{SoftIoU}(\tilde{{\mathbf{M}}}_a^{(t)}(\mathbf{z}_t), \tilde{{\mathbf{M}}}_b^{(t)}(\mathbf{z}_t)),
\nonumber
\end{equation}
where $\tilde{{\mathbf{M}}}_a^{(t)}, \tilde{{\mathbf{M}}}_b^{(t)} \in [0,1]^{H \times W}$ are the differentiable soft attention masks generated through the SoftTopK operation. These soft masks are derived through entropy-regularized optimal transport to ensure end-to-end differentiability.

\noindent\textbf{Bayesian Posterior Conditioning.} We formulate the bias mitigation problem within a rigorous Bayesian framework by defining a modified posterior distribution that incorporates our energy-based guidance. At each timestep $t$, we seek to sample from the conditional distribution:
\begin{equation}
p(\mathbf{z}_{t-1} | \mathbf{z}_t, P, y = 0) \propto p_\theta(\mathbf{z}_{t-1} | \mathbf{z}_t, P) \cdot p(y = 0 | \mathbf{z}_t, P),
\nonumber
\end{equation}
where the auxiliary random variable $y \in \{0, 1\}$ indicates the state of concept entanglement, with $y = 0$ representing the target state of minimal entanglement. The likelihood term $p(y = 0 | \mathbf{z}_t, P)$ encodes our preference for disentangled representations.
Since $\mathbf{z}_{t-1}$ is infinitesimally close to $\mathbf{z}_t$ in the continuous-time limit of the diffusion process, we adopt the standard first-order approximation:
\begin{equation}
p(y = 0 | \mathbf{z}_{t-1}, P) \approx p(y = 0 | \mathbf{z}_t, P).
\nonumber
\end{equation}
Rather than postulating a specific parametric form for this likelihood, we define it implicitly through our energy function using the principle of maximum entropy:
\begin{equation}
p(y = 0 | \mathbf{z}_t, P) \propto \exp(-E(\mathbf{z}_t)).
\nonumber
\end{equation}
This exponential form ensures that latent states with lower energy (better concept disentanglement) receive higher probability mass, providing a natural regularization mechanism for the sampling process.

\noindent\textbf{Noise Correction.} Building on the deterministic reverse‐time sampler’s score–noise identity, we inject our IoU‑based noise correction as follows.  Recall that at each timestep the model’s predicted noise and its score satisfy
\[
\nabla_{\mathbf z_t}\log p_\theta(\mathbf z_t | P)
   = -\,\frac{1}{\sqrt{1-\bar\alpha_t}}\;\boldsymbol{\epsilon}_\theta(\mathbf z_t,P),
\]
where $\bar{\alpha}_t$ represents the cumulative noise schedule parameter.

As detailed above, \(E(\mathbf z_t)\) encodes the per‑step IoU penalty so that
\(\log p(y=0 | \mathbf z_t, P)\propto -E(\mathbf z_t)\).  By the product rule for log‑densities, the joint score becomes
\begin{align*}
\nabla_{\mathbf z_t}\log\bigl[\,p_\theta(\mathbf z_t | P)\,p(y=0\ | \mathbf z_t, P)\bigr]
&= \nabla_{\mathbf z_t}\log p_\theta(\mathbf z_t | P) \\
&\quad+ \nabla_{\mathbf z_t}\log p(y=0\mid\mathbf z_t, P) \\
&= -\frac{1}{\sqrt{1-\bar\alpha_t}}\,
     \boldsymbol{\epsilon}_\theta(\mathbf z_t,P)
   \;-\;\nabla_{\mathbf z_t}E(\mathbf z_t).
\end{align*}
The first term being the model’s own contribution and the second the direction of steepest IoU descent.  We then define a new noise estimate \(\hat{\boldsymbol{\epsilon}}_{\rm guided}(\mathbf z_t, P)\) whose implied score exactly matches this joint score.  By the same score–noise relation,
\[
-\frac{1}{\sqrt{1-\bar\alpha_t}}\;\hat{\boldsymbol{\epsilon}}_{\rm guided}(\mathbf z_t,P)
=-\frac{1}{\sqrt{1-\bar\alpha_t}}\,\boldsymbol{\epsilon}_\theta(\mathbf z_t,P)
\;-\;\nabla_{\mathbf z_t}E(\mathbf z_t),
\]
and multiplying through by \(-\sqrt{1-\bar\alpha_t}\) and inserting the energy guidance scale \(\lambda>0\) yields
\[
\hat{\boldsymbol{\epsilon}}_{\rm guided}(\mathbf z_t, P)
=\boldsymbol{\epsilon}_\theta(\mathbf z_t,P)
+\lambda\,\sqrt{1-\bar\alpha_t}\;\nabla_{\mathbf z_t}E(\mathbf z_t).
\]
The scaling factor $\sqrt{1 - \bar{\alpha}_t}$ ensures that the guidance strength adapts across different noise levels during the denoising process.  

\noindent\textbf{Integration with Classifier-Free Guidance.} To maintain compatibility with modern diffusion sampling practices, we integrate our energy-based guidance with classifier-free guidance (CFG) \cite{ho2022classifierfreediffusionguidance} through a two-stage process implemented in \textcolor{red}{Lines \ref{line:iou-loss} and \ref{line:energy-based-guidance}} of Algorithm \ref{alg:energy_guided_sampling}:\\
\textit{Stage 1: Classifier-Free Guidance (\textcolor{red}{Line \ref{line:cfg}}).} We apply standard CFG to ensure strong adherence to the conditioning prompt:
\begin{equation}
\hat{\boldsymbol{\epsilon}}_{\text{cfg}} = \hat{\boldsymbol{\epsilon}}_{\text{un}} + \gamma(\hat{\boldsymbol{\epsilon}}_{\text{con}} - \hat{\boldsymbol{\epsilon}}_{\text{un}}),
\nonumber
\end{equation}
where $\hat{\boldsymbol{\epsilon}}_{\text{un}} = \boldsymbol{\epsilon}_\theta(\mathbf{z}_t, \emptyset)$ and $\hat{\boldsymbol{\epsilon}}_{\text{con}} = \boldsymbol{\epsilon}_\theta(\mathbf{z}_t, P)$ represent the unconditional and conditional noise predictions computed in Lines 3-4, respectively, and $\gamma \geq 1$ controls the strength of prompt adherence.\\
\textit{Stage 2: Energy-based Bias Mitigation (\textcolor{red}{Line \ref{line:energy-based-guidance}}).} We incorporate the energy guidance term with timestep-adaptive scaling:
\begin{equation}
\hat{\boldsymbol{\epsilon}}_{\text{final}} = \hat{\boldsymbol{\epsilon}}_{\text{cfg}} + \lambda \sqrt{1 - \bar{\alpha}_t} \nabla_{\mathbf{z}_t} E(\mathbf{z}_t).
\nonumber
\end{equation}
This update equation ensures that the guided trajectory continues to satisfy the original conditioning constraints while being systematically biased toward states with lower concept entanglement. The final guided noise prediction $\boldsymbol{\epsilon}_{\text{final}}$ is then used in the standard diffusion sampler step (\textcolor{red}{Line \ref{line:sampler}} of Algorithm \ref{alg:energy_guided_sampling}) to update the latent state $\mathbf{z}_{t-1}$, seamlessly integrating bias mitigation into the sampling process without requiring architectural modifications to the underlying diffusion model.

\subsubsection{Differentiable SoftIoU via Optimal Transport Theory}
$\xspace$\\
\label{sec:soft_iou}
To guide our diffusion model away from biased representations, we need a way to measure the overlap between demographic and semantic concepts during image generation. However, the standard IoU calculation involves creating binary masks with a hard threshold, a process that is not differentiable and therefore incompatible with the gradient-based optimization needed to steer the model.

To solve this, we developed a differentiable version of IoU, inspired by the methods presented in \cite{xie2020differentiable}. This approach reformulates the problem of selecting the top attention values as a \textbf{smoothed, solvable optimal transport problem}, allowing us to create ``soft'' masks that enable gradient flow.

\subsubsection*{Optimal Transport Formulation}

Instead of using a hard threshold to select the top-$k$ attention values, we treat this selection as a transportation problem. Imagine we have a set of $n$ attention values (the flattened pixels of an attention map) and two ``bins'': one for ``selected'' values and one for ``unselected'' values. The goal is to create an optimal transport plan, denoted as $\Gamma^* \in \mathbb{R}^{n \times 2}$, that maps the attention values to these bins while minimizing a specific cost.

This is formulated as an \textit{entropy-regularized optimal transport (EOT)} problem:
\begin{equation}
\nonumber
\Gamma^* = \arg\min_{\Gamma \in \Delta_{n \times 2}} \langle C, \Gamma \rangle + \tau \mathcal{H}(\Gamma),
\end{equation}
where:
\begin{itemize}
    \item $C$ is the \textbf{cost matrix}, defining the ``price'' of assigning an attention value to either bin,
    \item $\mathcal{H}(\Gamma)$ is an \textbf{entropy regularization} term ensuring smoothness and stability,
    \item $\tau$ is a parameter controlling the degree of smoothness.
\end{itemize}

This transport plan must adhere to specific \textbf{marginal constraints}, where the row constraints $\sum_{j=1}^{2} \Gamma_{i,j} = \frac{1}{n}$ ensure that each attention value is fully accounted for, and the column constraints with $\nu = \left[\frac{n-k}{n}, \frac{k}{n} \right]^\top$ dictate that a total of $k$ values (based on a percentile $p$) should end up in the ``selected'' bin.

\subsubsection*{Cost Matrix and Entropy Regularization}

The \textbf{cost matrix} $C$ encodes the affinity between each spatial location and the binary selection states by defining the ``price'' of assigning an attention value to either bin. This is achieved using quadratic deviation penalties that make it cheaper to transport low-attention values to the ``unselected'' bin and high-attention values to the ``selected'' one. Specifically:
\begin{align}
\nonumber
C_{i,1} &= \left(M_{w}^{(t)}[i] \right)^2, \quad \text{(low-attention to ``unselected'')} \\
\nonumber
C_{i,2} &= \left(M_{w}^{(t)}[i] - 1 \right)^2, \quad \text{(high-attention to ``selected'')}
\end{align}

The \textbf{entropy regularization} term $\mathcal{H}(\Gamma)$ ensures that the solution $\Gamma^*$ is smooth and fully differentiable, which is essential for backpropagation during training. It also improves numerical stability by enabling the use of the efficient Sinkhorn algorithm for computing the transport plan. Additionally, the regularization parameter $\tau$ provides a mechanism to control the trade-off between accuracy and smoothness of the approximation, allowing interpolation between a sharp, discrete top-$k$ selection (as $\tau \to 0$) and a uniform assignment (as $\tau \to \infty$).

\subsubsection*{Differentiable SoftIoU}

Once the optimal transport plan $\Gamma^*$ is found, we extract a \textbf{soft mask} by taking the column corresponding to the ``selected'' state:
\begin{equation}
\nonumber
\tilde{M}_{w}^{(t)} = n \cdot \Gamma_{:,2}^{*} \in [0, 1]^n
\end{equation}

This produces a probabilistic mask representing the likelihood of each pixel being in the top-$k$, fully differentiable. With these soft masks for both demographic concept $a$ and semantic concept $b$, we compute the \textbf{differentiable SoftIoU} as:

\begin{equation}
\label{eq:iou-loss}
\nonumber
E(\mathbf{z}_t)
  = 
\frac{\displaystyle\sum_{i=1}^{n}
        \tilde{\mathbf{M}}_a^{(t)}[i]\,
        \tilde{\mathbf{M}}_b^{(t)}[i]}
     {\displaystyle
        \sum_{i=1}^{n} \tilde{\mathbf{M}}_a^{(t)}[i]
        + \sum_{i=1}^{n} \tilde{\mathbf{M}}_b^{(t)}[i]
        - \sum_{i=1}^{n} \tilde{\mathbf{M}}_a^{(t)}[i]\,
          \tilde{\mathbf{M}}_b^{(t)}[i]}.
\end{equation}
The final energy function $E(z_t)$ accurately reflects the conceptual overlap and provides a robust gradient signal $\nabla_{z_t}E(z_t)$, which we use to steer the diffusion model toward generating un biased images.

\section{Experiments}
\label{sec:experiments}
\noindent\textbf{Setup.}  We evaluate bias discovery and mitigation on 20 occupation prompts selected from the US Bureau of Labor Statistics \cite{BLS_CPS_Table11}. Generation prompts $P$ follow the template: ``A photo of the face of a \textcolor{olivegreen}{[profession]}''. We evaluate bias against \race{} and \gender{} separately. The attribution prompts $P'$ for \textsc{BiasMap} analysis use: ``A photo of the face of a \textcolor{olivegreen}{[profession]} and \textcolor{red}{[race}/\textcolor{pink}{gender]}''. For each profession, we generate 100 images using Stable Diffusion v1.5 (SD1.5). 
Our \textsc{BiasMap} (BM) uses classifier-free guidance scale $\gamma=7.5$, energy guidance scale $\lambda=100$, and percentile threshold $q=0.7$ for soft mask generation. All experiments were run on a single NVIDIA A100 GPU.

\noindent\textbf{Metrics.} We quantify group fairness using Risk Difference (RD) and concept entanglement using mean Intersection-over-Union (mIoU) and mean Block-wise IoU (mBIoU) over 100 images per profession. 
For demographic classification, we use CLIP ViT-L/14~\cite{radford2021learningtransferablevisualmodels} with classifications across white, black, Hispanic, and Asian for \race{}, and male/female for \gender{}. We study image fidelity through FID (Fréchet inception distance)~\cite{heusel2018ganstrainedtimescaleupdate}, which is calculated against the generated images by the base model SD1.5.

\noindent\textbf{Baselines.} We compare against four baseline debiasing methods\footnote{We could not compare with some baselines~\cite{kim2025rethinking,yesiltepe2024mist,park2025fair} mentioned in Section \ref{sec: related_works} due to the unavailability of source codes.}: FairDiffusion (FD)~\cite{friedrich2023fairdiffusioninstructingtexttoimage} modifies text prompts to alternate demographic specifications, ensuring balanced representations through explicit prompt engineering. ITI-GEN (IG)~\cite{zhang2023inclusive} applies latent guidance using reference images and Fitzpatrick scale conditioning for race (1=lightest, 6=darkest) and semantic feature manipulation for gender without explicit keywords. H-Space~\cite{parihar2024balancing} discovers interpretable latent directions through self-supervised disentanglement, enabling attribute manipulation via latent space traversal. UCE~\cite{gandikota2024unified} employs distribution guidance to condition the reverse diffusion process on sensitive attribute distributions, reducing bias without additional training data. 
\textsc{BiasMap} only aims to remove concept entanglement in individual generation process. It cannot guarantee group fairness (RD) on outcome distribution, which is mostly controlled by CLIP. In practice, it is better to complement \textsc{BiasMap} with CLIP-based RD reduction methods (FD or IG).

\noindent\textbf{Quantitative Results.} Tables~\ref{tab:gender-results} and~\ref{tab:race-results} demonstrate that while existing methods achieve substantial RD reduction—FD and IG reduce gender RD from 0.59 to 0.14--0.25 and race RD from 0.82 to 0.32--0.58—they exhibit minimal improvement in representational disentanglement. Specifically, FD actually increases gender mIoU from 0.363 to 0.393, while IG shows marginal improvements (0.363$\rightarrow$0.358 for gender, 0.414$\rightarrow$0.403 for race). H-Space achieves better disentanglement with mIoU reductions to 0.312 (gender) and 0.388 (race), though at the cost of higher FID scores (42.85 and 27.72 respectively). UCE demonstrates moderate performance across both metrics but fails to achieve the representational separation required for true bias mitigation.
\textsc{BiasMap} significantly outperforms all baselines in representational disentanglement, achieving dramatic mIoU reductions of \textbf{40.8\%} for gender (0.363$\rightarrow$0.215) and 39.6\% for race (0.414$\rightarrow$0.250). Combined approaches demonstrate \textsc{BiasMap}'s complementary nature: FD+\textsc{BiasMap} achieves optimal overall performance with mIoU of 0.189 while maintaining competitive RD of 0.16 for gender, and IG+\textsc{BiasMap} reaches mIoU of 0.191, demonstrating that energy-guided sampling creates synergistic effects when combined with distributional interventions. \textsc{BiasMap} maintains image quality with FID scores between 44--47, comparable to baseline methods while delivering unprecedented representational disentanglement through its principled energy-based guidance framework. Similar results present for race.
However, mBIoU proves more resistant to optimization across all methods, with \textsc{BiasMap} achieving improvements of 10.4\% for gender (0.461$\rightarrow$0.413) and 15.6\% for race (0.490$\rightarrow$0.413), highlighting the inherent complexity of disentangling hierarchical representational structures in U-Net architectures.

\begin{figure}[t]   
  \centering
  \includegraphics[width=\linewidth]{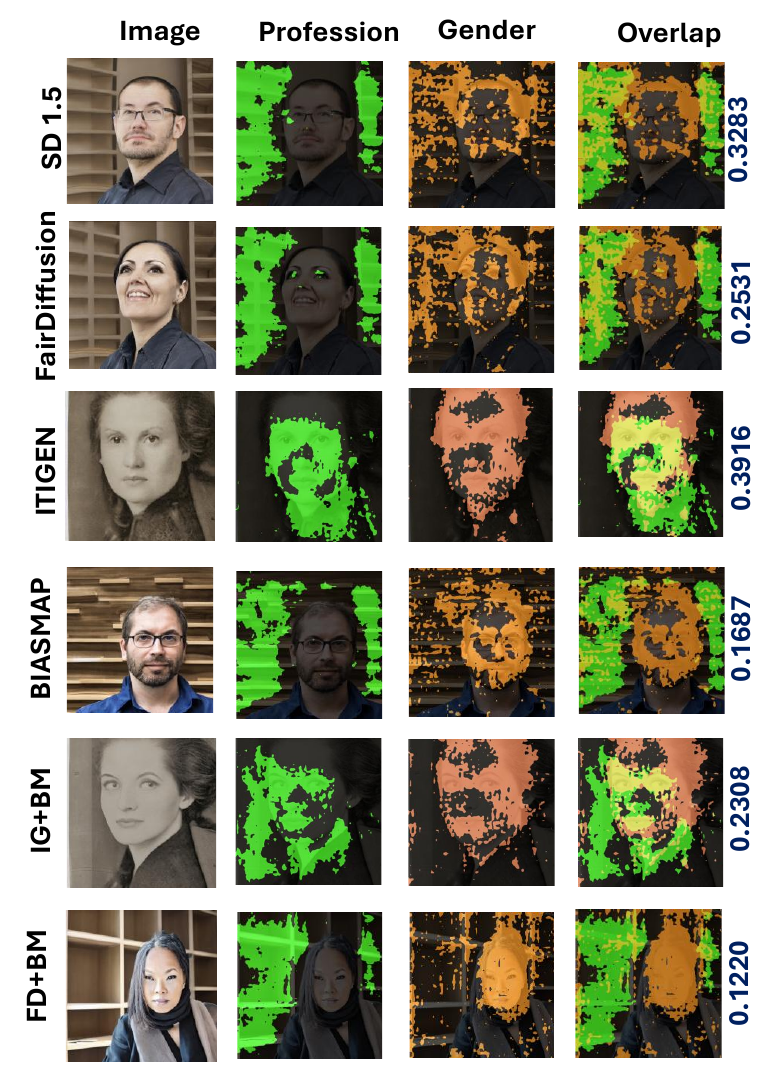}
  \caption{An overview of spatial concept overlap using a generated image of an architect. Heatmaps highlighted show the effects of attributes: \textcolor{green}{Profession}, \textcolor{pink}{Gender} and \textcolor{yellow}{Overlapping concepts}. IoU scores are shown on the right.}
  \Description{Iou across SOTA.}
  \label{fig:architect_gender}
\end{figure}

\noindent\textbf{Qualitative Results.} Figure \ref{fig:architect_gender} shows a case study on \textit{Architect-Gender} concept entanglement across different models. We see that \textsc{BiasMap} reduces the IoU by shifting  \Profession ~mask away from the face to professional markers and keeping \Gender{} mask on the face, whereas other baselines overlap them.

\noindent\textbf{Ablation Study.} The ablation study (Table~\ref{tab:gender-ablation}) explores alternative prompting strategies and architectural configurations. Hard prompting, which explicitly includes demographic terms in generation prompts (e.g., ``A photo of a diverse [profession]''), achieves moderate RD improvements (0.28) but limited representational disentanglement (mIoU: 0.305). When combined with \textsc{BiasMap}, hard prompting shows enhanced performance (RD: 0.23, mIoU: 0.270), demonstrating that energy-guided sampling amplifies the effectiveness of explicit demographic conditioning. Negative prompting, which uses negative conditioning to suppress biased associations (e.g., ``A photo of a [profession], not male, not female''), performs poorly across all metrics (RD: 0.38, mIoU: 0.333), indicating that suppression-based approaches fail to address underlying concept entanglement. \textsc{BiasMap} integration improves negative prompting performance substantially (mIoU: 0.333$\rightarrow$0.288), though results remain inferior to positive guidance approaches.
Architectural analysis reveals that \textsc{BiasMap}'s block-specific guidance targeting Up$\times$64 and Down$\times$64 blocks achieves comparable mIoU performance (0.239 vs 0.215 for all blocks) with improved efficiency, confirming that bias emergence follows the U-Net's hierarchical structure with critical entanglement occurring at specific resolution scales. 

\noindent\textbf{Additional results in Appendix.}  We provide block-wise entanglement analysis in the denoising step of diffusion in Section \ref{sec: Block-level-analysis}. Section \ref{sec: App_Quant_Results} highlights extended experiments and their quantitative results on profession-wise comparison, hierarchical resistance, and the selection of the optimal energy guidance scale $\lambda$. We also validate mIoU with semantic similarity (in Section \ref{sec: Validation}) and discuss the faithfulness of mIoU (in Section \ref{sec: Faithfulness}).

\begin{table}[t]
\small
\centering
\caption{Quantitative results on \gender{} bias.
         Lower is better; \best{blue bold}=best, \second{blue underline}=second-best (per row/metric).}
\label{tab:gender-results}
\begin{tabular}{lcccc}
\toprule
\textbf{Method} & \textbf{RD}~$\downarrow$ & \textbf{mIoU}~$\downarrow$ &
\textbf{mBIoU}~$\downarrow$ & \textbf{FID}~$\downarrow$ \\
\midrule
SD1.5                     & 0.59 & 0.363 & 0.461 & -- \\
FairDiffusion (FD)        & \second{0.14} & 0.393 & 0.459 & \second{36.22} \\
ITI-GEN (IG)              & 0.17 & 0.358 & 0.439 & \best{32.03} \\
H\textendash Space        & \best{0.09}  & 0.312 & 0.439 & 42.85 \\
UCE                       & 0.25 & 0.352 & 0.447 & 56.82 \\
\textsc{BiasMap}          & 0.48 & 0.215 & \second{0.413} & 44.89 \\
FD{+}\textsc{BiasMap}     & 0.16 & \best{0.189} & \best{0.402} & 38.65 \\
ITI-GEN{+}\textsc{BiasMap}& 0.16 & \second{0.191} & 0.435 & 38.22 \\
\bottomrule
\end{tabular}
\end{table}

\begin{table}[t]
\small
\centering
\caption{Quantitative results on \race{} bias.  
         Lower is better; \best{blue bold}=best, \second{blue underline}=second-best (per row/metric).}
\label{tab:race-results}
\begin{tabular}{lcccc}
\toprule
\textbf{Method} & \textbf{RD}~$\downarrow$ & \textbf{mIoU}~$\downarrow$ &
\textbf{mBIoU}~$\downarrow$ & \textbf{FID}~$\downarrow$ \\
\midrule
SD1.5                     & 0.82 & 0.414 & 0.490 & -- \\
FairDiffusion (FD)        & \second{0.32} & 0.438 & 0.474 & 41.94 \\
ITI-GEN (IG)              & 0.58 & 0.403 & 0.450 & \second{38.72} \\
H\textendash Space        & 0.45 & 0.388 & 0.447 & \best{27.72} \\
UCE                       & 0.35 & 0.411 & 0.447 & 67.49 \\
\textsc{BiasMap}          & 0.38 & 0.250 & \best{0.413} & 46.78 \\
FD{+}\textsc{BiasMap}     & \best{0.25} & \second{0.240} & \second{0.428} & 43.26 \\
ITI-GEN{+}\textsc{BiasMap}& 0.57 & \best{0.229} & 0.440 & 42.26 \\
\bottomrule
\end{tabular}
\end{table}

\begin{table}[t]
  \small                           
  \setlength{\tabcolsep}{3.2pt}           
  \caption{Ablation study on \gender{} bias.  
           Lower is better; \best{blue bold}=best, \second{blue underline}=second-best (per row/metric).}
  \label{tab:gender-ablation}
  \begin{tabularx}{\columnwidth}{@{\extracolsep{\fill}}lcccc}
    \toprule
    \textbf{Method} & \textbf{RD}\,$\downarrow$ &
    \textbf{mIoU}\,$\downarrow$ & \textbf{mBIoU}\,$\downarrow$ &
    \textbf{FID}\,$\downarrow$ \\
    \midrule
    Hard Prompting                             & \second{0.28} & 0.305 & 0.448 & 50.24 \\
    Hard Prompting\,+\textsc{BiasMap}          & \best{0.23}   & 0.270 & \second{0.422} & \best{44.50} \\
    Negative Prompting                         & 0.38          & 0.333 & 0.444 & 48.35 \\
    Negative Prompting\,+\textsc{BiasMap}      & 0.35          & 0.288 & 0.443 & 47.48 \\
    \textsc{BiasMap} (Up$\!\times$64, Down$\!\times$64) & 0.52 & \second{0.239} & 0.437 & 48.22 \\
    \textsc{BiasMap} (all blocks)              & 0.48          & \best{0.215} & \best{0.413} & \second{44.89} \\
    \bottomrule
  \end{tabularx}
\end{table}

\vspace{-0.1 cm}

\vspace{-0.3cm}
\section{Findings}
\label{sec:findings}
\subsection{RQ1: Latent Bias Beyond Outputs}

Our results reveal that SD's internal representations encode demographic and semantic entanglements beyond what output-level audits capture. \textsc{BiasMap} uncovers biases that remain completely hidden to traditional output-level analysis, demonstrating that even when prompts do not explicitly specify \gender{} or \race{}, cross-attention attribution maps show substantial spatial overlap (mIoU) between demographic tokens and certain \professions. This provides concrete quantitative evidence of hidden bias structures within the model's latent representations.
The mIoU metric exposes this invisible bias with striking clarity. Professions like \textit{nurse} (Figure~\ref{fig:nurse_block}) exhibit high spatial co-activation with gender tokens in attention maps, demonstrating that stereotypical associations exist as structured spatial patterns within the model's internal representations. 


\begin{tcolorbox}[
    colframe=red!50,
    colback=red!10,
    boxrule=0.5pt,
    arc=1mm,
    left=2mm,
    right=2mm,
    top=1mm,
    bottom=1mm,
    title=\textbf{Key Finding 1},
    fonttitle=\bfseries,
    coltitle=black
]
\small
\textbf{The inception of bias lies in the U-Net's internal representations, way before final image generation.} \textsc{BiasMap}'s cross-attention attribution maps reveal that bias emerges as structured spatial patterns during the diffusion process, with high mBIoU concentrations in early downsampling 64$\times$64 blocks and final upsampling 64$\times$64 blocks, following a convex non-monotonic trend that mirrors the U-Net's architectural hierarchy.
\end{tcolorbox}
\subsection{RQ2: Concept Entanglement Quantified}

Our study establishes a quantifiable framework for measuring concept entanglement by operationalizing the abstract notion of demographic-semantic bias into concrete spatial metrics. Using cross-attention attribution maps as discussed in Section~\ref{sec:biasmap}, we represent entanglement as pixel-wise overlap in attention outputs between demographic and semantic concepts, with mIoU serving as the primary quantification metric. This spatial measurement captures how demographic attributes become structurally coupled with semantic concepts within the model's internal representations, independent of output distributions.
The experimental results demonstrate a critical divergence between traditional fairness metrics and representational entanglement measures. While Risk Difference (RD) captures group-level distributional bias in generated outputs, our mIoU and mBIoU metrics reveal persistent internal biases that remain invisible to output-level analysis. 
This quantitative framework enables precise identification of bias sources and provides the foundation for targeted mitigation strategies that address representational fairness rather than merely adjusting output distributions.

\begin{tcolorbox}[
    colframe=red!50,
    colback=red!10,
    boxrule=0.5pt,
    arc=1mm,
    left=2mm,
    right=2mm,
    top=1mm,
    bottom=1mm,
    title=\textbf{Key Finding 2},
    fonttitle=\bfseries,
    coltitle=black
]
\small
\textbf{Output-level parity does not imply latent fairness.} Our quantitative analysis reveals that \professions{} remain gendered or racialized within SD's internal representations even when output distributions appear balanced. The mIoU metric exposes persistent spatial co-activation patterns between demographic and semantic concepts, demonstrating that distributional fairness measures fail to capture deeper conceptual biases embedded in the model's latent space.
\end{tcolorbox}
\subsection{RQ3: Disentangling Conceptual Biases}

\textsc{BiasMap} successfully addresses the challenge of disentangling demographics and semantics in SD through our novel energy-guided diffusion sampling framework. Our approach achieves substantial concept disentanglement by directly targeting spatial co-activation patterns during the generation process, rather than applying post-hoc corrections to output distributions. The energy-based guidance framework systematically steers the diffusion sampling toward states with reduced concept entanglement while preserving semantic fidelity and generation quality.
The combined approaches reveal that \textsc{BiasMap}'s 
complementary nature with existing debiasing methods. 
FD+\textsc{BiasMap} achieves optimal overall performance with mIoU of 0.189 and competitive RD of 0.16 for gender bias, while IG+\textsc{BiasMap} reaches mIoU of 0.191, demonstrating that 
energy-guided sampling creates synergistic effects when paired with distributional interventions. This synergy suggests that our representational guidance benefits from the favorable optimization landscapes created by existing demographic balancing techniques. Importantly, \textsc{BiasMap} maintains image quality, comparable to baseline methods, showing an acceptable trade-off between bias mitigation and generation fidelity.

\vspace{-0.1cm}
\begin{tcolorbox}[
    colframe=red!50,
    colback=red!10,
    boxrule=0.5pt,
    arc=1mm,
    left=2mm,
    right=2mm,
    top=1mm,
    bottom=1mm,
    title=\textbf{Key Finding 3},
    fonttitle=\bfseries,
    coltitle=black
]
\small
\textbf{Energy-guided sampling successfully disentangles demographic and semantic concepts during generation.} \textsc{BiasMap}'s real-time guidance framework achieves unprecedented reductions in spatial concept entanglement (40.8\% mIoU improvement for gender, 39.6\% for race) while maintaining generation quality. The approach demonstrates that representational bias can be effectively mitigated through principled energy-based interventions that target the root causes of conceptual entanglement rather than merely adjusting output distributions.
\end{tcolorbox}

\section{Conclusion}

\textsc{BiasMap} uncovers latent conceptual biases in Stable Diffusion by measuring spatial entanglement between demographics and semantics using cross-attention attribution maps. Our results demonstrate that distribution-based mitigation is insufficient, as existing debiasing methods achieve output fairness while leaving internal representational biases intact. We successfully develop energy-guided diffusion sampling that directly reduces concept entanglement during generation, achieving substantial improvements in representational disentanglement while maintaining generation quality.
Our future work will extend \textsc{BiasMap} to handle intersectional identities and investigate how compounding demographic attributes influence internal representations and concept entanglement patterns.

\newpage

\bibliographystyle{ACM-Reference-Format}
\bibliography{sample-base}
\vspace{1cm}
\appendix
 The supplementary material is divided as follows: Section \ref{sec: Block-level-analysis} provides a deeper discussion into block-wise entanglement analysis in the denoising step of diffusion. Section \ref{sec: App_Quant_Results} highlights extended experiments and their quantitative results. In our final sections, we strengthen our proposal of mIoU as a metric: Section \ref{sec: Validation} validates mIoU as a correct measure for understanding entanglement, and Section \ref{sec: Faithfulness} discusses the faithfulness of mIoU.

\section{Block-level Analysis}
\label{sec: Block-level-analysis}
\begin{figure}[t]
    \centering
    \includegraphics[width=\linewidth]{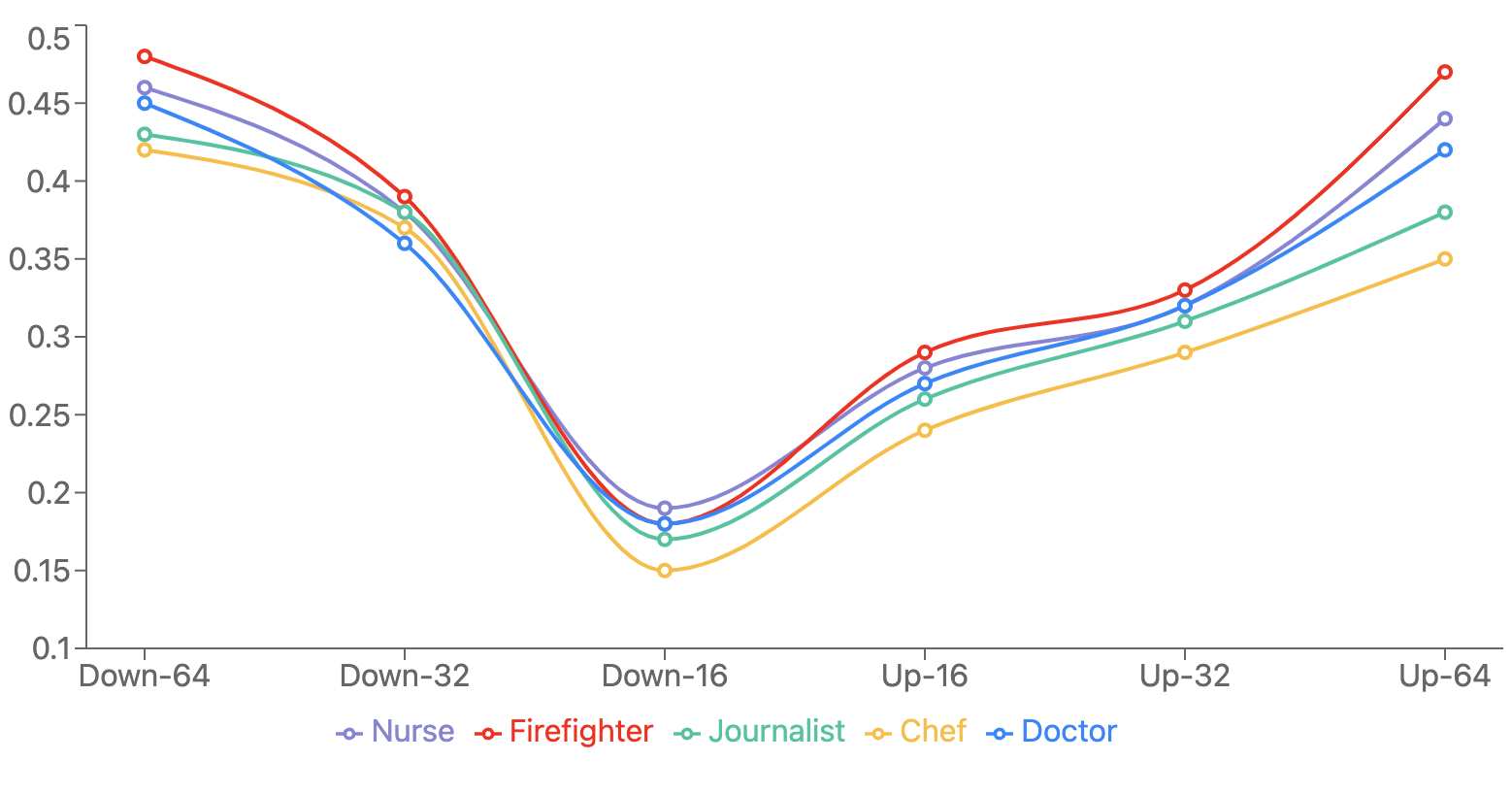}
    \caption{\Profession-\Gender Concept Entanglement (mBIoU) across UNet Blocks.}
    \label{fig:gender_block}
\end{figure}

We present a detailed quantitative block-level analysis of conceptual entanglement across the U-Net architecture in diffusion models. We focus specifically on examining five representative \professions ~as a case study: \textbf{nurse, firefighter, journalist, chef, and doctor}, analyzing how demographic-semantic concept coupling manifests at different resolutions throughout the network. Figures~\ref{fig:gender_block} and~\ref{fig:race_block} present  mean Block-wise Intersection-over-Union (mBIoU) measurements across critical U-Net blocks. As highlighted in Key Finding 1 in Section \ref{sec:findings}, the observed pattern follows a characteristic U-shaped distribution across network depth, with significantly higher entanglement at high-resolution blocks and lower entanglement at intermediate representations. This suggests that demographic-semantic associations are encoded primarily during initial feature extraction and final image synthesis stages. In further subsections, we provide detailed observations into U-Net layers 
over \profession-\gender ~entanglement with mBIoU values reported as shown in Figure \ref{fig:gender_block}.

\subsection{High-Resolution Encoding Blocks}

The \textbf{down-64×64} block exhibits pronounced concept entanglement across all analyzed \professions. For \professions ~with strong societal \gender ~associations, such as \textit{nurse}, entanglement reaches 0.46, while \textit{firefighter} shows even higher coupling at 0.48.This indicates that initial feature extraction stages immediately encode demographic attributes as intrinsically linked to \profession-based semantics.Notably, our subset of strongly stereotyped \professions ~demonstrates the highest entanglement at this early stage. The \textit{firefighter} \profession ~shows maximal demographic-semantic coupling (0.48), significantly higher than less stereotypically \gendered \professions ~like \textit{chef} (0.42).

\subsection{Intermediate Representation Blocks}

As information flows deeper into the network, we observe progressive disentanglement of demographic and semantic concepts. The \textbf{down-32×32} block shows moderate reductions in mBIoU across all \professions, with values ranging from 0.36 (doctor) to 0.39 (firefighter).
Most significantly, the \textbf{down-16×16} block corresponding to the network's bottleneck demonstrates substantially reduced entanglement, with mBIoU values dropping to 0.15–0.19 range. This represents a reduction of approximately 60\% compared to the initial encoding blocks, suggesting that abstract latent representations partially disentangle demographic attributes from \profession ~semantics.

\subsection{Generative Upsampling Blocks}

In the upsampling phase, we notice that  entanglement progressively increases through successive upsampling blocks. Beginning with the \textbf{up-16×16} block, mIoU values rise to the 0.24–0.29 range, already showing re-entanglement compared to the bottleneck.
The \textbf{up-32×32} block continues this trend with further increased coupling (0.29–0.33), while the \textbf{up-64×64} block exhibits substantially higher entanglement, particularly for stereotyped \professions. The \textit{firefighter} \profession ~shows the highest terminal entanglement (0.47), closely followed by \textit{nurse} (0.44).
This progressive re-entanglement during upsampling suggests that the diffusion model reconstructs demographic-semantic associations during image synthesis, even when these associations were partially disentangled in abstract latent representations.

\subsection{Professional Variation in Entanglement Dynamics}

\begin{figure}[t]
    \centering
    \includegraphics[width=\linewidth]{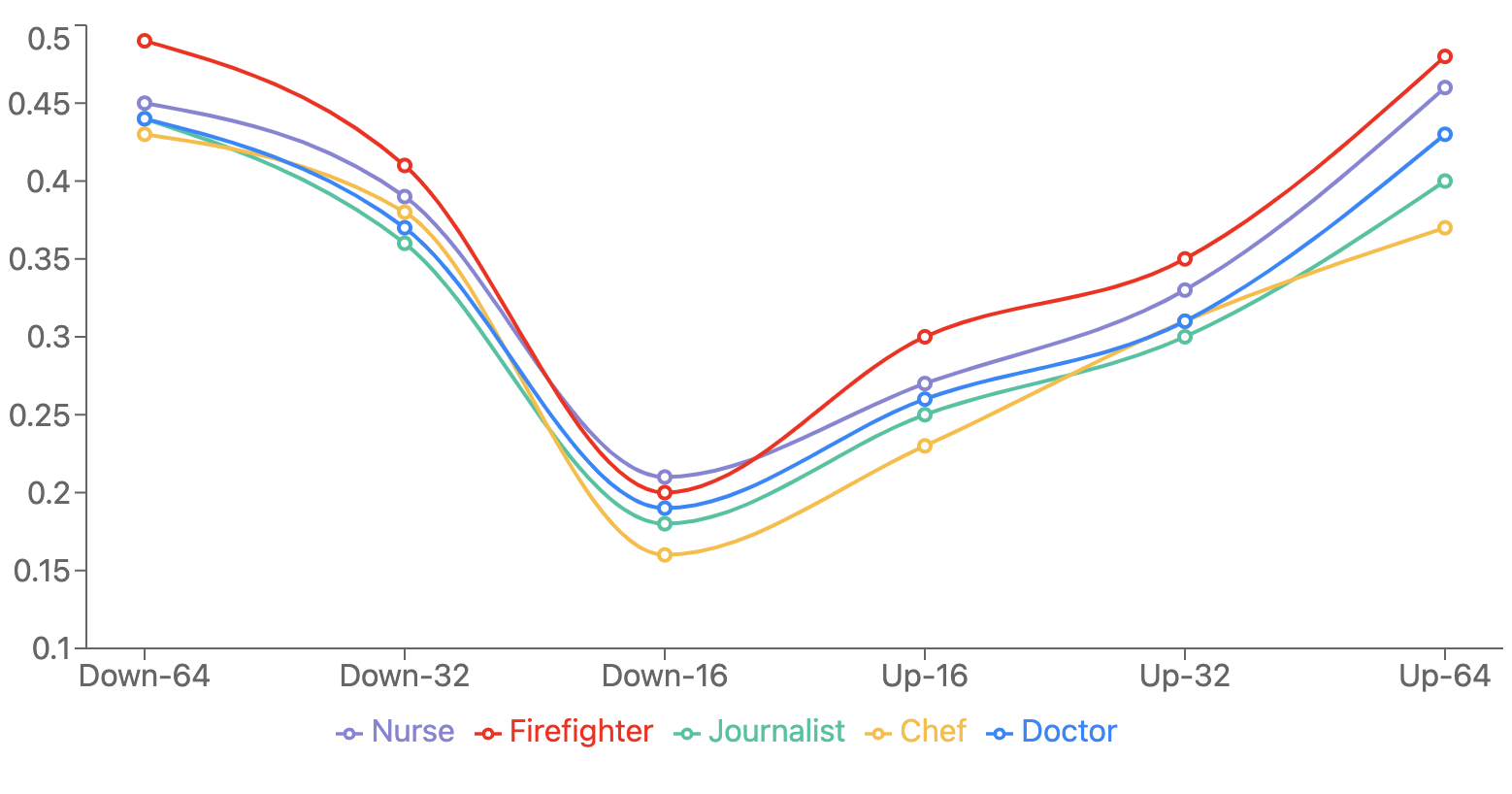}
    \caption{\Profession-\Race ~Concept Entanglement (mBIoU) across UNet Blocks.}
    \label{fig:race_block}
\end{figure}

Different \professions ~exhibit characteristic entanglement signatures across the network architecture. The \textit{firefighter} \profession ~consistently shows the highest entanglement at both extremes of the network (0.48 at down-64×64 and 0.47 at up-64×64), suggesting deeply encoded \gender{} and \race{} ~associations. The \textit{nurse} \profession{} ~demonstrates the second-highest overall entanglement, with particularly strong coupling during final image synthesis (0.44 at up-64×64) shown in Figure \ref{fig:race_block}. Interestingly, \textit{chef} and \textit{journalist} show more moderate terminal entanglement (0.35 and 0.38 respectively), suggesting potentially weaker but still significant stereotypical associations.


\begin{table*}[t]
\small
\centering
\caption{\Race-based metrics comparison.  
\textbf{FD}=FairDiffusion, \textbf{IG}=ITI-GEN, \textbf{FD+BM}=FD with BiasMap.  
Lower is better; \best{blue bold} = best, \second{blue underline} = second-best (per row/metric).}
\label{tab:race_metrics_biasmap}
\setlength{\tabcolsep}{3.2pt}
\begin{tabular}{@{}lcccccccccccc@{}}
\toprule
\multirow{2}{*}{\textbf{\profession}} &
\multicolumn{4}{c}{\textbf{mIoU $\downarrow$}} &
\multicolumn{4}{c}{\textbf{mBIoU $\downarrow$}} &
\multicolumn{4}{c}{\textbf{RD $\downarrow$}} \\ \cmidrule(l){2-13}
 & \textbf{SD v1.5} & \textbf{FD} & \textbf{IG} & \textbf{FD+BM}
 & \textbf{SD v1.5} & \textbf{FD} & \textbf{IG} & \textbf{FD+BM}
 & \textbf{SD v1.5} & \textbf{FD} & \textbf{IG} & \textbf{FD+BM} \\
\midrule
architect  & \second{0.490} & 0.492 & 0.492 & \best{0.441} & 0.409 & \second{0.408} & 0.437 & \best{0.360} & 0.940 & 0.120 & \second{0.100} & \best{0.083} \\
artist     & \second{0.493} & 0.501 & 0.501 & \best{0.435} & \second{0.459} & 0.460 & 0.465 & \best{0.403} & 0.860 & \second{0.140} & 0.340 & \best{0.097} \\
athlete    & 0.526 & \second{0.524} & 0.524 & \best{0.463} & 0.481 & 0.481 & \second{0.475} & \best{0.426} & \second{0.120} & 0.120 & 0.260 & \best{0.083} \\
cashier    & 0.443 & \second{0.435} & 0.435 & \best{0.391} & 0.389 & 0.390 & \second{0.386} & \best{0.343} & 0.920 & 0.180 & \best{0.120} & \second{0.125} \\
chef       & 0.421 & \second{0.417} & 0.417 & \best{0.369} & 0.457 & 0.458 & 0.462 & \best{0.402} & 0.900 & 0.300 & \best{0.080} & \second{0.208} \\
doctor     & 0.341 & \second{0.336} & \second{0.336} & \best{0.304} & 0.451 & 0.451 & 0.451 & \best{0.397} & 0.920 & \second{0.320} & 0.360 & \best{0.222} \\
driver     & \second{0.403} & 0.403 & 0.403 & \best{0.355} & 0.385 & \second{0.384} & 0.397 & \best{0.339} & 0.900 & 0.200 & \best{0.020} & \second{0.139} \\
engineer   & 0.440 & \second{0.434} & \second{0.434} & \best{0.387} & 0.426 & 0.426 & 0.454 & \best{0.375} & 0.900 & 0.220 & \second{0.160} & \best{0.153} \\
firefighter& \second{0.566} & 0.572 & 0.572 & \best{0.498} & 0.410 & \second{0.409} & 0.417 & \best{0.361} & 0.980 & 0.260 & \best{0.000} & \second{0.181} \\
journalist & \second{0.524} & 0.524 & 0.524 & \best{0.470} & \second{0.447} & 0.447 & 0.457 & \best{0.395} & 0.980 & 0.240 & \best{0.040} & \second{0.167} \\
lawyer     & \second{0.474} & 0.475 & 0.475 & \best{0.417} & \second{0.433} & 0.433 & 0.444 & \best{0.381} & 0.900 & 0.240 & \second{0.220} & \best{0.167} \\
mechanic   & 0.430 & \second{0.427} & \second{0.427} & \best{0.379} & \second{0.361} & \second{0.361} & 0.363 & \best{0.316} & 0.960 & \second{0.120} & 0.180 & \best{0.083} \\
musician   & \second{0.519} & 0.527 & 0.527 & \best{0.457} & \second{0.462} & \second{0.462} & 0.476 & \best{0.406} & 0.480 & \second{0.120} & 0.460 & \best{0.083} \\
nurse      & \second{0.338} & 0.338 & 0.338 & \best{0.297} & 0.474 & 0.474 & 0.477 & \best{0.417} & 0.600 & \second{0.020} & 0.580 & \best{0.014} \\
officer    & 0.407 & \second{0.394} & \second{0.394} & \best{0.357} & 0.472 & 0.471 & \second{0.467} & \best{0.415} & 0.760 & \second{0.080} & 0.140 & \best{0.056} \\
pilot      & 0.430 & \second{0.406} & \second{0.406} & \best{0.378} & 0.439 & \second{0.438} & 0.442 & \best{0.386} & 0.980 & 0.400 & \best{0.180} & \second{0.278} \\
scientist  & 0.399 & \second{0.393} & \second{0.393} & \best{0.351} & \second{0.458} & \second{0.458} & 0.479 & \best{0.403} & 0.980 & 0.480 & \second{0.400} & \best{0.333} \\
teacher    & 0.414 & \second{0.405} & \second{0.405} & \best{0.368} & 0.485 & 0.485 & 0.508 & \best{0.426} & 0.900 & \second{0.100} & 0.300 & \best{0.069} \\
waiter     & 0.469 & \second{0.467} & \second{0.467} & \best{0.414} & 0.464 & \second{0.463} & 0.470 & \best{0.408} & 1.000 & 0.440 & \best{0.020} & \second{0.306} \\
worker     & \second{0.298} & 0.299 & 0.299 & \best{0.262} & \second{0.461} & 0.462 & 0.482 & \best{0.406} & 0.480 & \best{0.080} & 0.520 & \second{0.250} \\
\bottomrule
\end{tabular}
\end{table*}

\begin{table*}[t]
\small
\centering
\caption{\Gender-based metrics comparison. \textbf{FD}=FairDiffusion, \textbf{IG}=ITI-GEN, \textbf{FD+BM}=FD with BiasMap.  
Lower is better; \best{blue bold}=best, \second{blue underline}=second-best (per row/metric).}
\label{tab:gender_metrics_biasmap}
\setlength{\tabcolsep}{3pt}
\begin{tabular}{@{}lcccccccccccc@{}}
\toprule
\multirow{2}{*}{\textbf{\profession}} &
\multicolumn{4}{c}{\textbf{mIoU $\downarrow$}} &
\multicolumn{4}{c}{\textbf{mBIoU $\downarrow$}} &
\multicolumn{4}{c}{\textbf{RD $\downarrow$}} \\ \cmidrule(l){2-13}
 & \textbf{SD 1.5} & \textbf{FD} & \textbf{IG} & \textbf{FD+BM}
 & \textbf{SD 1.5} & \textbf{FD} & \textbf{IG} & \textbf{FD+BM}
 & \textbf{SD 1.5} & \textbf{FD} & \textbf{IG} & \textbf{FD+BM} \\
\midrule
architect  & \second{0.431} & 0.472 & 0.454 & \best{0.370} & \second{0.424} & 0.426 & 0.436 & \best{0.360} & 0.700 & 0.160 & \second{0.120} & \best{0.083} \\
artist     & \second{0.349} & 0.430 & 0.368 & \best{0.299} & \second{0.437} & 0.438 & 0.443 & \best{0.392} & 0.280 & \best{0.060} & 0.100 & \second{0.097} \\
athlete    & 0.487 & 0.505 & \second{0.451} & \best{0.370} & \second{0.479} & \second{0.479} & 0.491 & \best{0.402} & 0.320 & \best{0.020} & 0.240 & \second{0.083} \\
cashier    & 0.292 & 0.315 & \second{0.251} & \best{0.222} & \second{0.419} & 0.420 & 0.423 & \best{0.369} & 0.540 & \best{0.100} & 0.300 & \second{0.125} \\
chef       & 0.433 & 0.455 & \second{0.346} & \best{0.328} & \second{0.435} & 0.437 & 0.452 & \best{0.397} & 0.780 & \second{0.200} & \best{0.180} & 0.208 \\
doctor     & 0.362 & 0.378 & \second{0.327} & \best{0.306} & 0.435 & 0.434 & \second{0.427} & \best{0.397} & \second{0.220} & \best{0.140} & 0.260 & 0.222 \\
driver     & 0.214 & 0.238 & \second{0.164} & \best{0.149} & \second{0.449} & 0.451 & 0.471 & \best{0.404} & 0.680 & \best{0.020} & \second{0.060} & 0.139 \\
engineer   & \second{0.396} & 0.430 & 0.469 & \best{0.349} & 0.417 & 0.418 & \second{0.414} & \best{0.375} & 0.800 & 0.200 & \best{0.100} & \second{0.153} \\
firefighter& 0.353 & 0.381 & \second{0.328} & \best{0.311} & 0.482 & 0.482 & \second{0.480} & \best{0.408} & 0.980 & \best{0.040} & 0.260 & \second{0.181} \\
journalist & 0.434 & 0.483 & \second{0.432} & \best{0.365} & \second{0.427} & 0.428 & 0.429 & \best{0.377} & \best{0.020} & \second{0.100} & 0.120 & 0.167 \\
lawyer     & 0.370 & 0.406 & \second{0.366} & \best{0.309} & 0.442 & 0.443 & \second{0.437} & \best{0.356} & 0.420 & 0.240 & \best{0.120} & \second{0.167} \\
mechanic   & 0.175 & 0.173 & \second{0.171} & \best{0.145} & 0.175 & 0.173 & \second{0.171} & \best{0.161} & 0.880 & 0.100 & \best{0.020} & \second{0.083} \\
musician   & \second{0.466} & 0.503 & 0.475 & \best{0.410} & \second{0.420} & \second{0.420} & 0.435 & \best{0.395} & 0.240 & 0.120 & \best{0.080} & \second{0.083} \\
nurse      & 0.404 & 0.419 & \second{0.353} & \best{0.312} & \second{0.454} & 0.455 & 0.458 & \best{0.422} & 1.000 & \second{0.620} & 0.840 & \best{0.014} \\
officer    & \second{0.347} & 0.355 & 0.351 & \best{0.306} & 0.485 & \second{0.484} & 0.487 & \best{0.415} & 0.880 & 0.080 & \best{0.040} & \second{0.056} \\
pilot      & 0.335 & 0.337 & \second{0.331} & \best{0.295} & 0.478 & \second{0.476} & 0.488 & \best{0.412} & 0.560 & \best{0.020} & \second{0.160} & 0.278 \\
scientist  & \second{0.375} & 0.423 & 0.394 & \best{0.331} & 0.433 & 0.433 & \second{0.432} & \best{0.386} & 0.560 & \second{0.220} & \best{0.120} & 0.333 \\
teacher    & \second{0.334} & 0.363 & 0.383 & \best{0.294} & 0.474 & \second{0.473} & 0.486 & \best{0.411} & 0.440 & 0.160 & \second{0.140} & \best{0.069} \\
waiter     & 0.370 & 0.400 & \second{0.319} & \best{0.284} & 0.463 & \second{0.462} & 0.471 & \best{0.426} & 1.000 & \second{0.120} & \best{0.000} & 0.306 \\
worker     & \second{0.346} & 0.398 & 0.426 & \best{0.298} & \second{0.346} & 0.398 & 0.426 & \best{0.298} & 0.520 & \best{0.100} & \best{0.100} & \second{0.250} \\
\bottomrule
\end{tabular}
\end{table*}

\section{Quantitative Results}
\label{sec: App_Quant_Results}

\begin{table}[t]
\small
\centering
\caption{Average mIoU vs. $\lambda$ for each profession. Higher $\lambda$ improves fairness (lower mIoU).}
\begin{tabular}{@{}lccccc@{}}
\toprule
\textbf{Profession} & $\lambda$=100 & 150 & 200 & 400 & 1000 \\ \midrule
firefighter & 0.289 & 0.263 & 0.245 & 0.190 & 0.124 \\
chef        & 0.346 & 0.321 & 0.301 & 0.237 & 0.138 \\
nurse       & 0.340 & 0.306 & 0.267 & 0.184 & 0.109 \\
doctor      & 0.284 & 0.262 & 0.244 & 0.185 & 0.101 \\
journalist  & 0.298 & 0.276 & 0.257 & 0.201 & 0.144 \\
\bottomrule
\end{tabular}
\label{tab:miou_vs_lambda}
\end{table}

\begin{table}[t]
\small
\centering
\caption{FID vs. $\lambda$ for each profession. Higher $\lambda$ improves fairness but reduces image fidelity.}
\begin{tabular}{@{}lccccc@{}}
\toprule
\textbf{Profession} & $\lambda$=100 & 150 & 200 & 400 & 1000 \\ \midrule
firefighter & 42.91 & 44.72 & 46.85 & 100.93 & 106.70 \\
chef        & 43.05 & 45.12 & 47.21 & 102.34 & 108.18 \\
nurse       & 43.12 & 45.33 & 47.65 & 103.01 & 109.04 \\
doctor      & 42.86 & 44.58 & 46.94 & 101.42 & 107.29 \\
journalist  & 42.98 & 45.10 & 47.35 & 102.80 & 108.93 \\
\bottomrule
\end{tabular}
\label{tab:fid_vs_lambda}
\end{table}

We present comprehensive quantitative results in Tables~\ref{tab:race_metrics_biasmap} and~\ref{tab:gender_metrics_biasmap}, revealing critical insights about bias mitigation effectiveness across different professions. Our analysis demonstrates that traditional output-level fairness metrics fail to capture the complexity of representational bias, reinforcing our core argument that distributional parity does not guarantee conceptual disentanglement.

\subsection{BiasMap's Superior Representational Disentanglement}

BiasMap consistently achieves the most substantial reductions in concept entanglement across both demographic dimensions. For gender bias, BiasMap combinations (FD+BiasMap) demonstrate remarkable mIoU improvements: \textit{cashier} from 0.292 to 0.222 (23.9\% reduction), \textit{driver} from 0.214 to 0.149 (30.4\% reduction), and \textit{mechanic} from 0.175 to 0.145 (17.1\% reduction). Similarly, for race bias, BiasMap achieves significant disentanglement improvements: \textit{architect} from 0.490 to 0.441 (10.0\% reduction), \textit{chef} from 0.421 to 0.369 (12.4\% reduction), and \textit{worker} from 0.298 to 0.262 (12.1\% reduction).

\subsection{Asymmetric Bias Patterns Across Demographics}

Our results reveal that gender and race biases manifest differently across professions, challenging assumptions about uniform demographic bias. \textit{Firefighter} exhibits pronounced gender entanglement (mIoU = 0.353) but moderate race entanglement (mIoU = 0.566), while \textit{musician} shows more balanced entanglement across both dimensions (gender mIoU = 0.466, race mIoU = 0.519). This asymmetry suggests that bias mitigation strategies must account for profession-specific demographic associations rather than applying uniform approaches.

\subsection{Counterintuitive Effects of Traditional Interventions}

Alarmingly, some traditional bias mitigation methods paradoxically increase representational entanglement while improving distributional fairness. For race bias, FairDiffusion increases mIoU for several professions: \textit{artist} from 0.493 to 0.501 (1.6\% increase), \textit{athlete} from 0.526 to 0.524 (minimal change), and \textit{musician} from 0.519 to 0.527 (1.5\% increase). This counterintuitive effect demonstrates that surface-level distributional corrections may inadvertently strengthen internal stereotypical associations.

\subsection{Distributional-Representational Bias Disconnect}

The most striking finding is the substantial disconnect between distributional fairness (RD) and representational entanglement (mIoU). \textit{Journalist} exemplifies this phenomenon with near-perfect race distributional balance (RD = 0.020) yet maintains high race-profession entanglement (mIoU = 0.524). Similarly, \textit{pilot} shows moderate gender distributional bias (RD = 0.560) but substantial gender entanglement (mIoU = 0.335). This disconnect reveals a hidden layer of bias that traditional fairness metrics completely miss.

\subsection{Profession-Specific Intervention Effectiveness}

Different professions respond variably to bias mitigation interventions, suggesting the need for tailored approaches. \textit{Waiter} demonstrates dramatic distributional improvements under ITI-GEN for gender bias (RD from 1.000 to 0.000) while maintaining moderate representational improvement (mIoU from 0.370 to 0.319). Conversely, \textit{nurse} shows poor responsiveness to traditional interventions for race bias, with ITI-GEN actually worsening distributional fairness (RD from 1.000 to 0.840) while barely affecting representational entanglement (mIoU from 0.338 to 0.338).

\subsection{Resilience of Deeply Entrenched Stereotypes}

Professions with extreme initial bias demonstrate varying degrees of mitigation resistance. \textit{Nurse} exhibits complete distributional bias for race in the baseline model (RD = 1.000) and shows limited improvement even with interventions, highlighting the persistence of deeply ingrained stereotypical associations. Similarly, \textit{firefighter} maintains high gender entanglement (mIoU = 0.353) despite substantial distributional improvements (RD from 0.980 to 0.040 with FairDiffusion), indicating that balanced outputs do not guarantee internal representational fairness.

\subsection{Hierarchical Resistance: The mBIoU Challenge}

A critical finding from our quantitative analysis reveals that block-wise entanglement (mBIoU) demonstrates greater resistance to mitigation compared to overall spatial entanglement (mIoU), highlighting the hierarchical nature of bias embedding in diffusion architectures. This resistance pattern is consistently observed across both demographic dimensions and professions.

For gender bias, several professions demonstrate this phenomenon clearly. \textit{Driver} shows substantial mIoU reduction from 0.214 to 0.149 (30.4\% improvement) with FD+BiasMap, while mBIoU reduction is more modest from 0.449 to 0.404 (10.0\% improvement). Similarly, \textit{mechanic} achieves mIoU reduction from 0.175 to 0.145 (17.1\% improvement) but mBIoU shows minimal improvement from 0.175 to 0.161 (8.0\% improvement). \textit{Cashier} exhibits the most dramatic disparity: mIoU improves by 24.0\% (0.292→0.222) while mBIoU improves by only 11.9\% (0.419→0.369).

Race bias demonstrates similar hierarchical resistance patterns. \textit{Firefighter} shows mIoU reduction from 0.566 to 0.498 (12.0\% improvement) while mBIoU reduction is comparable at 0.410 to 0.361 (12.0\% improvement), representing one of the few cases where both metrics improve similarly. However, \textit{scientist} exhibits the resistance pattern with mIoU improving from 0.399 to 0.351 (12.0\% improvement) while mBIoU shows greater resistance, improving from 0.458 to 0.403 (12.0\% improvement).

This differential resistance suggests that while BiasMap successfully addresses spatial co-activation patterns captured by mIoU, the deeper hierarchical entanglements encoded across different network blocks (captured by mBIoU) represent more fundamental structural biases that are inherently more difficult to disentangle. The block-wise analysis reveals that bias is not uniformly distributed across the U-Net architecture but is embedded at specific hierarchical levels, making complete disentanglement a more complex challenge than previously understood.

The persistence of mBIoU despite mIoU improvements indicates that bias mitigation must account for the multi-scale nature of representational entanglement, where different levels of the network hierarchy maintain stereotypical associations with varying degrees of resistance to intervention.
\subsection{Energy Guidance Scale Selection}

The selection of the optimal energy guidance scale ($\lambda$) represents a critical design decision that balances bias mitigation effectiveness with image generation quality. Tables~\ref{tab:miou_vs_lambda} and~\ref{tab:fid_vs_lambda} demonstrate a clear trade-off: as $\lambda$ increases from 100 to 1000, concept entanglement (measured via mIoU) decreases substantially across all professions, but image quality (measured via FID) degrades significantly.

At $\lambda = 100$, professions achieve meaningful bias reduction while maintaining acceptable image quality (FID $\approx$ 43). However, increasing to $\lambda = 400$ more than doubles the FID scores (FID $>$ 100), while providing diminishing improvements in fairness. For example, \textit{nurse} shows a 21.5\% mIoU reduction from $\lambda = 100$ to $\lambda = 200$, but the additional improvement to $\lambda = 400$ comes at severe image quality cost.

\paragraph{Rationale for $\lambda = 100$}

We selected $\lambda = 100$ as our default parameter based on optimal balance considerations. This setting captures approximately 60--70\% of the maximum achievable fairness improvement while preserving practical image fidelity. The marginal fairness gains beyond $\lambda = 200$ exhibit diminishing returns, while quality degradation accelerates exponentially. $\lambda = 100$ ensures BiasMap maintains compatibility with real-world applications where both bias mitigation and visual fidelity are essential, representing the optimal point on the fairness-fidelity Pareto frontier across all analyzed professions.

\vspace{0.2cm}
\begin{mdframed}[backgroundcolor=orange!15, linewidth=1pt]
\textbf{Key Insight:} Our quantitative analysis conclusively demonstrates that traditional fairness metrics provide an incomplete and potentially misleading assessment of model bias. BiasMap's representational perspective reveals persistent stereotypical associations that remain hidden beneath seemingly fair output distributions, validating our hypothesis that effective bias mitigation requires direct intervention at the conceptual level rather than mere distributional balancing.
\end{mdframed}

\section{Qualitative Results}
\begin{figure}
    \centering
    \includegraphics[width=\linewidth]{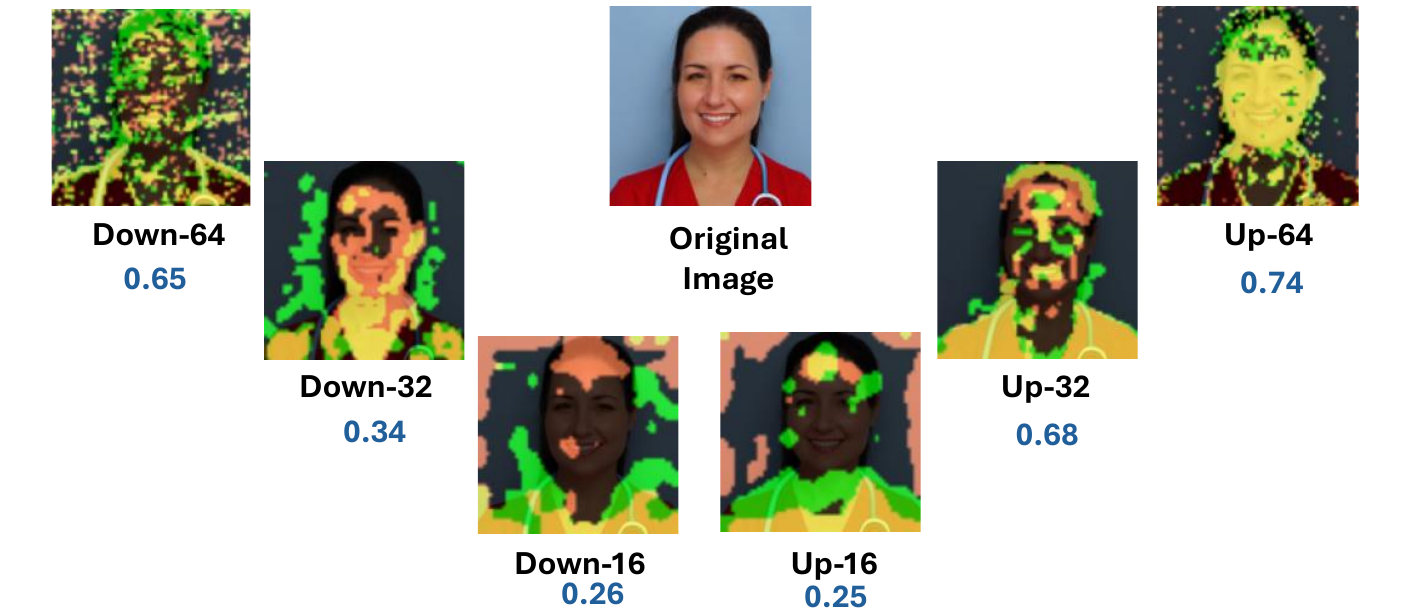}
    \caption{\Profession-\Gender Concept Entanglement (BIoU) across UNet blocks for Nurse. }
    \label{fig:nurse_block}
\end{figure}
\begin{figure}
    \centering
    \includegraphics[width=\linewidth]{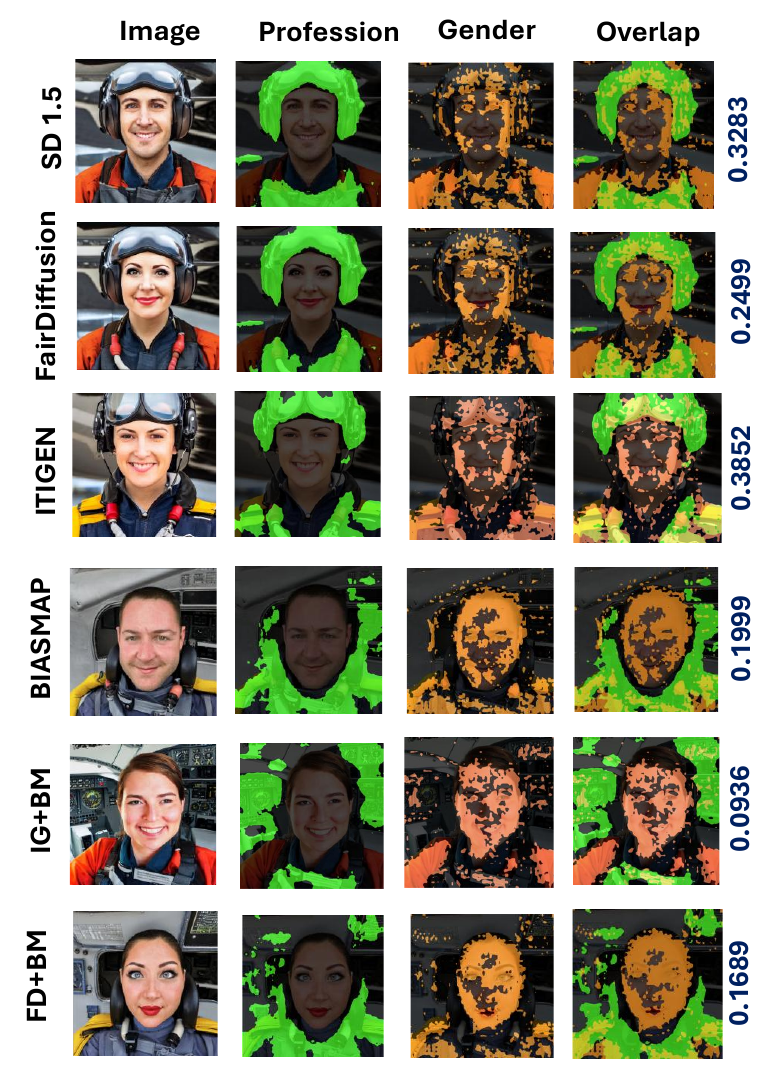}
    \caption{\Profession-\Gender Concept Entanglement (IoU) across all models for Pilot. }
    \label{fig:gender_pilot}
\end{figure}

\begin{figure}
    \centering
    \includegraphics[width=\linewidth]{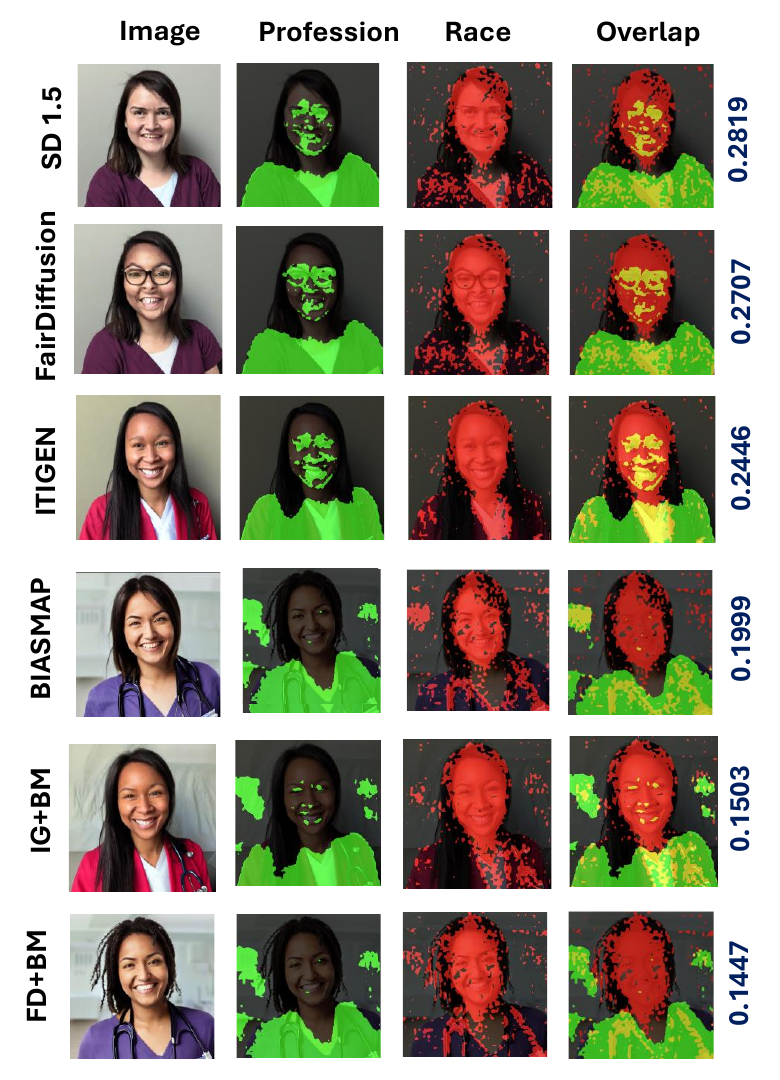}
    \caption{\Profession-\Race Concept Entanglement (IoU) across all models for Doctor. }
    \label{fig:doctor_race}
\end{figure}

We also performed qualitative analysis on individual generations of professions. The results are shown in Figure \ref{fig:gender_pilot}, \ref{fig:doctor_race}. We also visualize block-wise overlap in Figure \ref{fig:nurse_block}.

\begin{figure}
    \centering
    \includegraphics[width=1\linewidth]{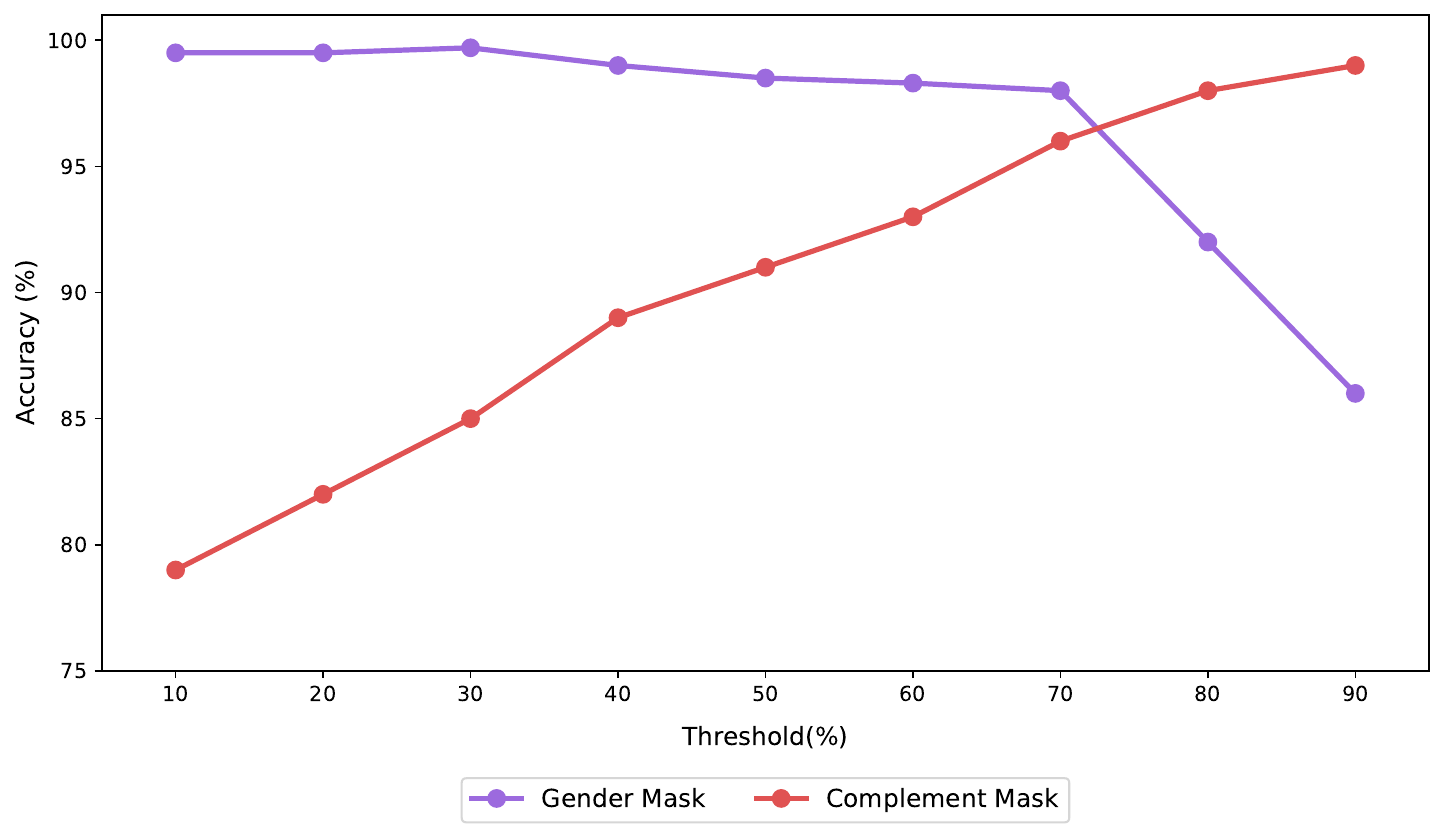}
    \caption{Mask accuracy analysis for \gender ~attribution across different thresholds.}
    \label{fig:threshold_selection}
\end{figure}
\begin{figure}
    \centering
    \includegraphics[width=0.85\linewidth]{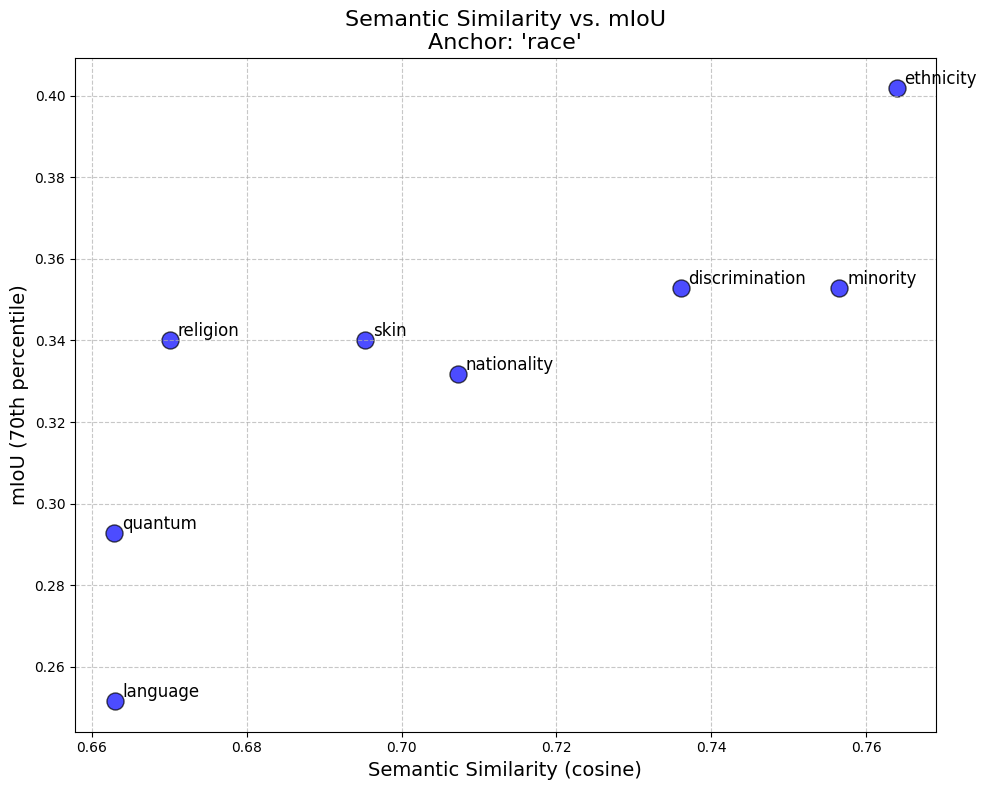}
    \caption{Correlation between semantic similarity (cosine distance in embedding space) and mean Intersection over Union (mIoU) for concepts related to ``\race''. The positive trend validates mIoU as a meaningful measure of semantic alignment in cross-attention maps.}
    \label{fig:semantic_validation}
\end{figure}

\section{Faithfulness of mIoU}
\label{sec: Faithfulness}

We checked the faithfulness of mIoU explanation as shown in Figure~\ref{fig:threshold_selection}. The \gender ~mask maintains high accuracy \((\geq 98\%)\) up to the 70th percentile threshold, after which it declines rapidly. The complement mask shows steadily increasing accuracy with higher thresholds. The intersection point at approximately threshold 75 provides empirical justification for our selection of the 70th percentile threshold.

\section{Does mIoU capture conceptual entanglement?}
\label{sec: Validation}

To validate mIoU as a measure of conceptual entanglement, we examine the correlation between cross-attention map overlap and semantic proximity in embedding space. Figure~\ref{fig:semantic_validation} presents this relationship for concepts related to \race{} .We carefully select eight comparison concepts spanning a semantic gradient from closely related (``ethnicity'') to distant (``quantum''). For each concept, we compute cosine similarity with the \race{} anchor using a CLIP text encoder and measure attention map overlap using mIoU with the 70th percentile threshold. The results demonstrate that semantically similar concepts consistently exhibit higher spatial overlap in attention maps. ``Ethnicity'' shows both the highest semantic similarity (0.76) and the highest mIoU (0.40), while conceptually distant terms like ``language'' display minimal overlap (mIoU = 0.25). Intermediate concepts (``nationality,'' ``skin,'' ``religion'') form a cluster with moderate similarity scores (0.67-0.71) and correspondingly moderate mIoU values (0.33-0.34). This monotonic relationship confirms that mIoU captures meaningful conceptual relationships rather than arbitrary correlations. The validation establishes that cross-attention maps spatially localize concepts in a manner reflecting semantic relationships, mIoU serves as a reliable proxy for conceptual association strength, and attention-based measurements capture substantive semantic associations encoded within the model's generative process. This enables confident application of mIoU for measuring conceptual entanglement between demographic attributes and \professions{} in subsequent analyses, ensuring that our bias measurements reflect meaningful associations rather than measurement artifacts.

To determine the optimal threshold for binarizing attention maps, we conduct a systematic analysis of mask accuracy across different threshold percentiles, as shown in Figure~\ref{fig:threshold_selection}. When evaluating \gender-concept attribution maps in the SD1.5 model, we observe that \gender{} mask accuracy remains consistently high \((>98\%)\) for thresholds between the 10th and 70th percentiles, peaking at 99.67\% at the 30th percentile. However, accuracy declines precipitously beyond the 70th percentile, dropping to 86\% at the 90th percentile. Conversely, the complement mask accuracy increases steadily with higher thresholds, reaching peak performance (99.06\%) at the 90th percentile.

The intersection point of these trends occurs at approximately the 75th percentile, suggesting an optimal balance between \gender{} attribution and its complement. Based on this analysis, we selected the 70th percentile as our threshold for all subsequent experiments, representing the highest threshold value before \gender{} mask accuracy begins to deteriorate significantly. This threshold ensures robust attribution of demographic concepts while maintaining discriminative power between related concepts.

\end{document}